\documentclass[lettersize,journal]{IEEEtran}
\usepackage{amsmath}
\usepackage{amsfonts}
\usepackage{array}
\usepackage{algpseudocode}
\usepackage[caption=false,font=normalsize,labelfont=sf,textfont=sf]{subfig}
\usepackage{textcomp,todonotes}
\usepackage{stfloats}
\usepackage{url}
\usepackage{verbatim}
\usepackage{graphicx}
\usepackage[dvipsnames]{xcolor}
\usepackage{cite}
\usepackage{times}
\usepackage{epsfig}
\usepackage[compact]{titlesec}
\usepackage{amssymb}
\usepackage{booktabs}
\usepackage{colortbl}
\usepackage[vlined,ruled]{algorithm2e}
\usepackage{color}
\usepackage{soul}
\usepackage{multirow}
\usepackage{float}
\usepackage{balance}
\usepackage{makecell}
\usepackage{caption}
\usepackage{subcaption}
\usepackage{lipsum}
\usepackage{rotating}
\usepackage{adjustbox}
\usepackage{pdflscape}
\usepackage{siunitx}
\usepackage{lscape}
\usepackage{pifont}
\usepackage{blindtext}
\usepackage{tabu}
\usepackage[T1]{fontenc}
\usepackage[utf8]{inputenc}
\usepackage{flexisym}
\usepackage{forest}
\usepackage{longtable}
\usepackage[multiple]{footmisc}
\usepackage{ctable}
\usepackage{epstopdf}
\usepackage{fixfoot}
\usepackage{multicol}
\usepackage{tabularray}
\usepackage{tabularx}
\usepackage[most]{tcolorbox}
\usepackage{xspace}
\usepackage{academicons}
\usepackage{xr-hyper}
\usepackage{etoolbox}

\definecolor{iccvblue}{rgb}{0.21,0.49,0.74}
\usepackage[pagebackref,breaklinks,colorlinks,allcolors=iccvblue]{hyperref}

\usepackage[capitalize]{cleveref}
\crefname{section}{Sec.}{Secs.}
\Crefname{section}{Section}{Sections}
\Crefname{table}{Table}{Tables}
\crefname{table}{Tab.}{Tabs.}

\makeatletter
\newcommand{\linebreakand}{%
\end{@IEEEauthorhalign}
\hfill\mbox{}\par
\mbox{}\hfill\begin{@IEEEauthorhalign}
}
\makeatother

\newcommand{\xmark}{\ding{55}}%

\newcommand{\orcid}[1]{\href{https://orcid.org/#1}{\textcolor[HTML]{A6CE39}{\aiOrcid}}}

\makeatletter
\DeclareRobustCommand\onedot{\futurelet\@let@token\@onedot}
\def\@onedot{\ifx\@let@token.\else.\null\fi\xspace}
\def\eg{\emph{e.g}\onedot} 
\def\ie{\emph{i.e}\onedot}

\def\etal{\emph{et al}\onedot}
\makeatother

\begin{document}


\title{A Cycle Ride to HDR: Semantics Aware Self-Supervised Framework for Unpaired LDR-to-HDR Image Reconstruction}

\author{
Hrishav Bakul Barua$^{\star}$, \IEEEmembership{Member, IEEE},
Kalin Stefanov, \IEEEmembership{Member, IEEE},
Lemuel Lai En Che,
Abhinav Dhall, \IEEEmembership{Member, IEEE},
KokSheik Wong$^{\dagger}$, \IEEEmembership{Senior Member, IEEE},
Ganesh Krishnasamy, \IEEEmembership{Member, IEEE}

\thanks{Lemuel Lai En Che, K. Wong, and G. Krishnasamy are with School of Information Technology, Monash University Malaysia, Malaysia (e-mail: \{hrishav.barua, wong.koksheik, ganesh.krishnasamy\}@monash.edu, lemuel.lai03@gmail.com).}

\thanks{H. B. Barua, K. Stefanov, and A. Dhall are with Faculty of Information Technology, Monash University, Australia (e-mail: kalin.stefanov, abhinav.dhall\}@monash.edu).}

\thanks{H. B. Barua is also with the Robotics and Autonomous Systems Lab, TCS Research, Kolkata, India}.





\thanks{$^{\star}$This research is supported by the Global Excellence and Mobility Scholarship, Monash University (Australia and Malaysia).}

\thanks{$^{\dagger}$This research is supported, in part, by the E-Science fund under the project: \emph{Innovative High Dynamic Range Imaging - From Information Hiding to Its Applications} (Grant No. 01-02-10-SF0327).}
}


\maketitle

\begin{abstract}
Reconstruction of High Dynamic Range (HDR) from Low Dynamic Range (LDR) images is an important computer vision task.
There is a significant amount of research utilizing both conventional non-learning methods and modern data-driven approaches, focusing on using both single-exposed and multi-exposed LDR for HDR image reconstruction.
However, most current state-of-the-art methods require high-quality paired \{LDR;HDR\} datasets with limited literature use of unpaired datasets, that is, methods that learn the LDR~$\leftrightarrow$~HDR mapping between domains.
This paper proposes CycleHDR, a method that integrates self-supervision into a modified semantic- and cycle-consistent adversarial architecture that utilizes unpaired LDR and HDR datasets for training.
Our method introduces novel artifact- and exposure-aware generators to address visual artifact removal.
It also puts forward an encoder and loss to address semantic consistency, another underexplored topic.
CycleHDR is the first to use semantic and contextual awareness for the LDR~$\leftrightarrow$~HDR reconstruction task in a self-supervised setup.
The method achieves state-of-the-art performance across several benchmark datasets and reconstructs high-quality HDR images.
\end{abstract}

\begin{IEEEkeywords}
High Dynamic Range, Cycle-consistency, Self-supervision, Adversarial learning, Unpaired data.
\end{IEEEkeywords}

\begin{figure}[t]
\centering
\includegraphics[width=\columnwidth]{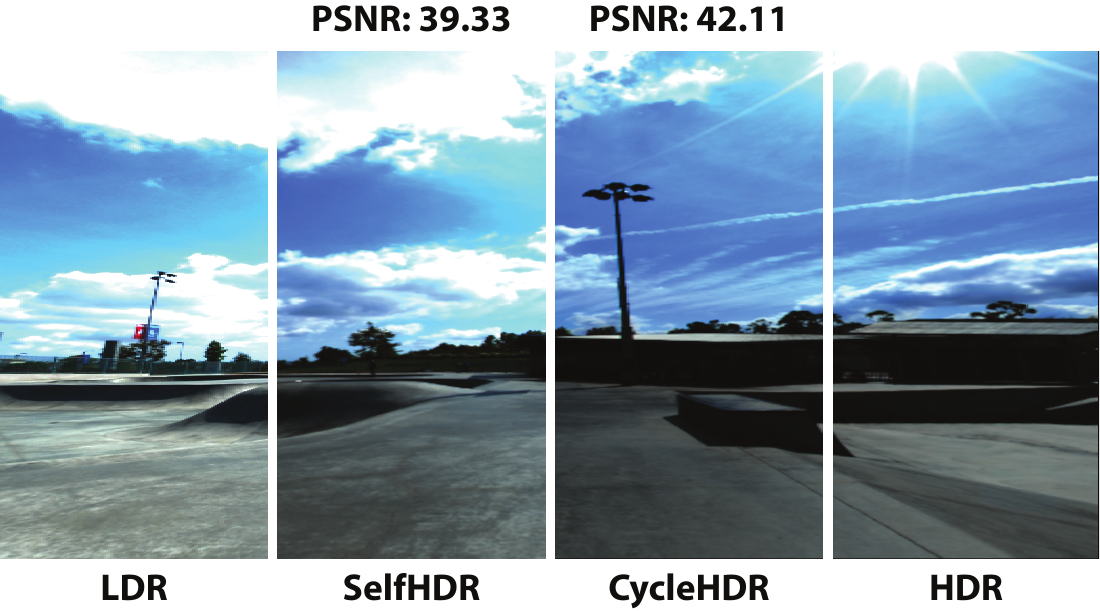}
\caption{Qualitative comparison of the proposed CycleHDR and state-of-the-art SelfHDR~\cite{SelfHDR} methods. CycleHDR handles the overexposed portions in the sky more realistically. Here, a single image is divided into four parts, each rendered using a different approach.}
\label{fig:teaser}
\end{figure}

\section{Introduction}
\label{sec:introduction}
High Dynamic Range (HDR)~\cite{artusi2019overview,mccann2011art} images capture a wider range of intensity values compared to their Low Dynamic Range (LDR) or Standard Dynamic Range (SDR) counterparts, which have a pixel bit depth of only $2^8$ intensity levels.
From human vision perspective, LDR images are often not visually pleasing, while from the computer/robot vision applications perspective, they hold limited information.

Most commercial devices produce and display LDR content, while some specialized hardware support HDR.
For example, HDR cameras and sensors~\cite{kalra2024towards} can capture images with more than 256 intensity levels, and HDR displays~\cite{cao2024perceptual} can render a range beyond 256 intensity levels. 
Given that 
HDR-capable devices are expensive, researchers have been investigating approaches for accurate HDR reconstruction from LDR (\ie, inverse tone-mapping~\cite{wang2021deep}) using conventional non-learning methods and modern data-driven approaches.
For display purposes, however, one often needs to tone-map~\cite{han2023high} the HDR content to LDR to fit the intensity range supported by standard hardware.

Early HDR reconstruction methods address tasks such as image enhancement, \eg, recovering missing information due to extreme lighting conditions~\cite{nguyen2023psenet} or low lighting conditions~\cite{yang2023lightingnet, LoLi} and compression~\cite{cao2024learned}.
As technology advances, HDR reconstruction expands its application domains, including robotics/machine vision~\cite{wu2020hdr,liu2025isethdr}, media and entertainment~\cite{he2022sdrtv}, gaming, mixed reality, and novel view synthesis~\cite{satilmis2023deep, lu2024pano, singh2024hdrsplat, huang2022hdr}, as well as medical imaging~\cite{huang2022hdr}.
Some of the earlier data-driven methods utilize Convolutional Neural Networks (CNN)~\cite{shin2018cnn}, Transformers~\cite{shang2024hdrtransdc,liu2022ghost} and Generative Adversarial Networks (GAN)~\cite{raipurkar2021hdr,niu2021hdr}.
Recent approaches for HDR image and/or 3D HDR scene reconstruction are based on Diffusion Models~\cite{dalal2023single, goswami2024semantic, bemana2024exposure, wang2024lediff}, Neural Radiance Fields (NeRF)~\cite{huang2022hdr}, and Gaussian Splatting~\cite{singh2024hdrsplat}.
Some use multi-exposed images~\cite{barua2023arthdr,santos2020single,eilertsen2017hdr,li2019hdrnet,liu2020single,le2023single,guo2023single,zou2023rawhdr,guo2022lhdr,ren2023robust}, while others use single-exposed LDR~\cite{barua2024histohdr,cai2018learning,SelfHDR} paired with HDR images for supervised training.

Although current methods achieve excellent results in reconstructing HDR images, most require proper \{LDR;HDR\} paired datasets~\cite{barua2024gta,liu2020single, endo2017deep} for training.
Consequently, the quality of the state-of-the-art data-driven HDR reconstruction methods depends on the quality of the available paired \{LDR;HDR\} datasets.
There is a research gap in the field; on the one hand, the literature on unpaired HDR reconstruction is extremely limited~\cite{li2022uphdr,wang2023glowgan}, and on the other, only a few approaches utilize semantic and contextual information to guide the reconstruction process~\cite{goswami2024semantic, goswami2022g, liu2022ghost, wang2022kunet}.

To address this gap, we propose \textbf{CycleHDR}, a method that leverages cycle consistency~\cite{zhu2017unpaired, chen2019mocycle} objective using unpaired data, where the model learns a mapping between domains, \ie, LDR~$\leftrightarrow$~HDR.
In addition, the method ensures semantic consistency between the LDR and reconstructed HDR by utilizing Contrastive Language-Image Pretraining (CLIP)~\cite{radford2021learning} embeddings and loss based on semantic segmentation.
Furthermore, a heuristic-based guidance of artifact and exposure information further supports the training process to deliver more realistic and natural HDR image reconstruction.
The proposed method reconstructs artifact-free and visually impressive HDR from single-exposed LDR images.
It also outperforms the most recent state-of-the-art, which predominantly uses paired datasets for training (see \cref{fig:teaser}).
Our work makes the following contributions:

\begin{itemize}
\item{We introduce the first semantic- and cycle consistency-guided self-supervised learning method for unpaired \{LDR;HDR\} data which addresses both the inverse tone-mapping (\ie, LDR~$\rightarrow$~HDR) and tone-mapping (\ie, HDR~$\rightarrow$~LDR) tasks (see \cref{sec:method})}.
\item{We leverage a simple heuristic-based guidance to obtain a loss function and artifact- and exposure-aware saliency maps to further refine those areas in the reconstructed HDR images (see \cref{subsec:architecture_modules} and \cref{subsec:loss_functions})}.
\item{We propose a novel generator based on a modified U-Net architecture~\cite{ronneberger2015u} that incorporates ConvLSTM-based artifact-aware feedback mechanism~\cite{huang2017densely,li2019feedback} and exposure-aware skip connections to mitigate visual artifacts in the HDR reconstruction (see \cref{subsec:architecture_modules})}.
\item{We propose a CLIP embedding encoder for contrastive learning to minimize the semantic difference between LDR and reconstructed HDR image pairs (see \cref{subsec:architecture_modules})}.
\item{We propose a novel loss function based on the Mean Intersection over Union (mIoU) metric to further ensure semantic consistency between the LDR and reconstructed HDR images (see \cref{subsec:loss_functions})}.
\item{We perform thorough experimental validation for the contribution of all proposed components, both qualitatively and quantitatively (see \cref{sec:results}).}
\end{itemize}

\section{Related Work}
\label{sec:related_work}
The most popular non-learning approach consists of multi-exposed LDR fusion to achieve high dynamic range in the output image~\cite{jinno2011multiple}.
This approach involves image feature alignment, calculating weights for feature mapping on the basis of image characteristics, and fusing the images on the basis of the weights to get the most appropriate exposures for each part of the image.
Another approach is using histogram equalization~\cite{dar2021enhanced} that allows for contrast and brightness levels re-adjustment in the over/underexposed regions of the image.
Gradient domain manipulation~\cite{kuk2011high} techniques enhance the granular details of an LDR image and expand its dynamic range.
Pyramid-based image fusion~\cite{keerativittayanun2015innovative} is a technique where image features are extracted into Laplacian pyramids~\cite{keerativittayanun2015innovative} and then fused for the HDR.
Some methods use Retinex-based~\cite{kim2011natural,meylan2006high} tuning to effectively produce and display HDR images on consumer screens~\cite{kim2011natural,meylan2006high}.
Intensity mapping function-based approaches~\cite{yao2011intensity, drago2003adaptive, qiao2015tone} use transformations on image pixels either by using logarithmic mapping or Gamma correction to stretch the intensity range or enhance lower intensity pixels.
The methods in this category, although time-efficient, face some issues including ghosting effects, halo artifacts, and blurring in the reconstruction, and they do not generalize well for variety of input LDR images.

\begin{table}[t]
\setlength{\tabcolsep}{2pt}
\small
\centering
\caption{Summary of recent state-of-the-art methods. \textbf{I/O:} LDR used as input, single-exposed (SE) and multi-exposed (ME), \textbf{O/P:} Reconstructs directly HDR (D) or multi-exposed LDR stack (I), \textbf{UP:} Can be trained with unpaired data, \textbf{HF:} Uses heuristic-based guidance of artifact and exposure information, \textbf{Con (Context):} Uses local/global image information and relationship among entities in the image, \textbf{Sem (Semantics):} Uses color/texture information and identity of the items in the image, \textbf{Art (Artifacts):} Handles visual artifacts in heavily over/underexposed areas, \textbf{TM}: Supports tone-mapping.}
\label{tab:related_work}
\begin{tabular}{lcccccccc}
\toprule[0.5mm]
\textbf{Method} & \textbf{I/O} & \textbf{O/P} & \textbf{UP} & \textbf{HF} & \textbf{Con} & \textbf{Sem} & \textbf{Art} & \textbf{TM} \\
\midrule[0.25mm]
PSENet~\cite{nguyen2023psenet} & SE & D & \textcolor{red}{\xmark} &\textcolor{red}{\xmark}& \textcolor{red}{\xmark} & \textcolor{red}{\xmark} & \textcolor{red}{\xmark} & \textcolor{red}{\xmark} \\
SingleHDR(W)~\cite{le2023single} & SE & I & \textcolor{red}{\xmark} &\textcolor{red}{\xmark}& \textcolor{red}{\xmark} & \textcolor{red}{\xmark} & \textcolor{red}{\xmark} & \textcolor{red}{\xmark} \\
UPHDR-GAN~\cite{li2022uphdr} & ME & D & \textcolor{Green}{\checkmark} &\textcolor{red}{\xmark} & \textcolor{red}{\xmark} & \textcolor{red}{\xmark} & \textcolor{Green}{\checkmark} & \textcolor{red}{\xmark} \\
SelfHDR~\cite{SelfHDR} & ME & I & \textcolor{red}{\xmark} &\textcolor{red}{\xmark}& \textcolor{red}{\xmark} & \textcolor{red}{\xmark} & \textcolor{Green}{\checkmark} & \textcolor{red}{\xmark} \\
KUNet~\cite{wang2022kunet} & SE & D & \textcolor{red}{\xmark} & \textcolor{red}{\xmark}&\textcolor{red}{\xmark} & \textcolor{Green}{\checkmark} & \textcolor{red}{\xmark} & \textcolor{red}{\xmark} \\
Ghost-free HDR~\cite{liu2022ghost} & ME & D & \textcolor{red}{\xmark} & \textcolor{red}{\xmark}&\textcolor{Green}{\checkmark} & \textcolor{red}{\xmark} & \textcolor{Green}{\checkmark} & \textcolor{red}{\xmark} \\
GlowGAN-ITM~\cite{wang2023glowgan} & SE & D & \textcolor{Green}{\checkmark} & \textcolor{red}{\xmark}& \textcolor{red}{\xmark} & \textcolor{red}{\xmark}  &\textcolor{Green}{\checkmark} & \textcolor{red}{\xmark} \\
DITMO~\cite{goswami2024semantic} & SE & I & \textcolor{red}{\xmark} &\textcolor{red}{\xmark} &\textcolor{red}{\xmark} & \textcolor{Green}{\checkmark} & \textcolor{Green}{\checkmark} & \textcolor{red}{\xmark} \\
\midrule[0.25mm]
\textbf{CycleHDR (Ours)} & \textbf{SE} & \textbf{D} & \textbf{\textcolor{Green}{\checkmark}} &\textcolor{Green}{\checkmark}& \textbf{\textcolor{Green}{\checkmark}}  &\textbf{\textcolor{Green}{\checkmark}} & \textbf{\textcolor{Green}{\checkmark}} & \textcolor{Green}{\checkmark} \\
\bottomrule[0.5mm]
\end{tabular}
\end{table}

Learning-based approaches use either single-exposed LDR as input~\cite{santos2020single,eilertsen2017hdr,li2019hdrnet,liu2020single,le2023single,guo2023single,zou2023rawhdr,guo2022lhdr} or alternatively, multi-exposed LDR~\cite{niu2021hdr,cai2018learning,ren2023robust,SelfHDR,barua2023arthdr} for HDR reconstruction.
Some of the methods that use multi-exposed LDR employ novel feature fusion or aggregation techniques~\cite{ye2021progressive, xiao2024deep, yan2023unified}.
Barua~\etal~\cite{barua2024histohdr} harnessed the concept of histogram equalization of input LDR, in addition to the original LDR, to overcome the contrast and hue miss-representation in over/underexposed areas of the LDR.
Cao~\etal~\cite{cao2023decoupled} proposed a channel-decoupled kernel-based approach which combines the output HDR with the output of another architecture in a pixel-wise fashion.
Luzardo~\etal~\cite{luzardo2020fully} proposed a method to enhance the artistic intent of the reconstructed HDR by enhancing the peak brightness in the reconstruction process.
NeRF~\cite{mildenhall2021nerf,huang2022hdr,mildenhall2022nerf,lu2024pano}, Diffusion Models~\cite{dalal2023single, goswami2024semantic,wang2024lediff}, and Gaussian Splatting~\cite{singh2024hdrsplat} have also been deployed to improve HDR reconstruction.

Some methods employ weakly-supervised~\cite{le2023single}, self-supervised~\cite{SelfHDR} and unsupervised~\cite{nguyen2023psenet, wang2023glowgan} approaches for HDR reconstruction.
Le~\etal~\cite{le2023single} proposed an indirect approach for HDR reconstruction from multi-exposed LDR using weak supervision.
The method first outputs a stack of multi-exposed LDR, which are then merged using the state-of-the-art tool Photomatix~\cite{photomatix} to obtain an HDR.
Zhang~\etal~\cite{SelfHDR} proposed a self-supervised approach for HDR reconstruction.
The method requires multi-exposed LDR to learn the reconstruction model without the need for HDR counterparts.
Nguyen~\etal~\cite{nguyen2023psenet} proposed an unsupervised approach to recover image information from overexposed areas of LDR.
This approach does not require any HDR for supervision and instead uses pseudo-ground truth images.
Lee~\etal~\cite{lee2020learning} proposed a method with limited supervision for multi-exposed LDR generation consisting of a pair of models that generate LDR with higher and lower exposure levels.
Then the two exposure levels are combined to generate the HDR.
Wang~\etal~\cite{wang2023glowgan} proposed an unsupervised approach for HDR reconstruction.
They train the model in such a way that when the HDR is re-projected to LDR using a camera response function model it becomes indistinguishable from the original LDR.

\noindent\textbf{GAN-Based Approaches:}
Generative Adversarial Networks~\cite{goodfellow2020generative, anand2021hdrvideo, wang2023glowgan} are well explored for HDR reconstruction.
Niu~\etal~\cite{niu2021hdr} proposed a GAN-based method that can handle images with foreground motions by fusing multi-exposed LDR and extracting important information from the over/underexposed areas of the LDR.
Raipurkar~\etal~\cite{raipurkar2021hdr} proposed a conditional GAN architecture that adds details to the saturated regions of the input LDR using a pre-trained segmentation model to extract exposure masks.
Guo~\etal~\cite{guo2023single} proposed a two-stage pipeline that extracts the over/underexposed features with high accuracy.
GAN with attention mechanism first generates the missing information in those extreme exposure areas, and in the second stage, a CNN with multiple branches fuses the multi-exposed LDR from the previous stage to reconstruct the HDR.
Nam~\etal~\cite{nam2024deep} proposed a GAN method that uses exposure values for conditional generation of multi-exposure stack that adapts well to varying color and brightness levels.

Li~\etal~\cite{li2022uphdr} proposed a GAN-based unsupervised method for unpaired multi-exposure LDR-to-HDR translation.
The method introduces modified GAN loss and a novel discriminator to tackle ghosting artifacts caused from the misalignment in the LDR stack and HDR.
Wu~\etal~\cite{wu2023cycle} proposed CycleGAN-like architecture for low-light image enhancement using unpaired data.
The method fuses the Retinex theory~\cite{land1977retinex} with CycleGAN~\cite{zhu2017unpaired} concept to enhance the lighting conditions globally, recover color and reduce noise in the output HDR.

\begin{figure*}[t]
\captionsetup[subfigure]{justification=centering}
\centering
\subfloat[Forward cycle consistency $x \rightarrow G_{Y}(x) \rightarrow G_{X}(G_{Y}(x)) \approx x$.]{\includegraphics[width=\columnwidth]{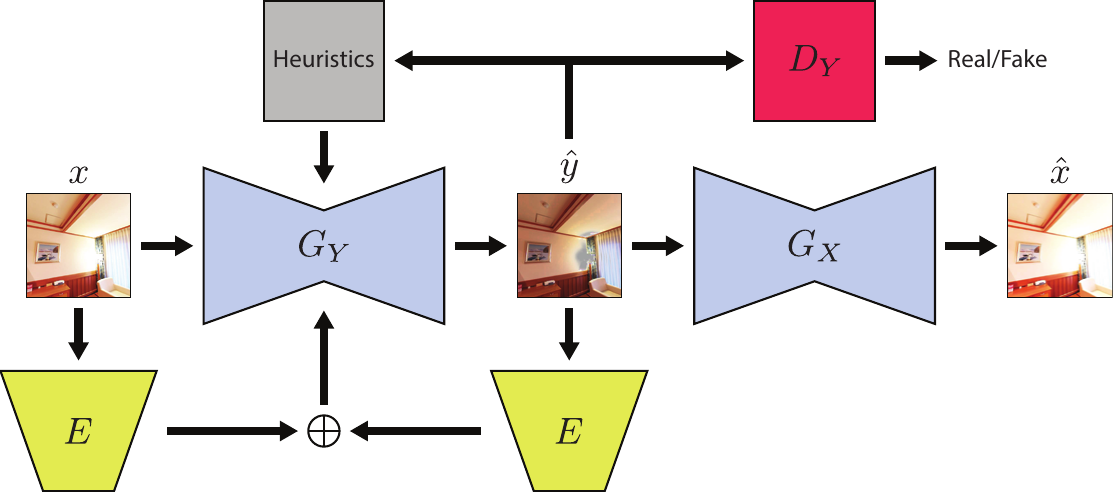}\label{subfig:method_forward}}
\hfill
\subfloat[Backward cycle consistency $y \rightarrow G_{X}(y) \rightarrow G_{Y}(G_{X}(y)) \approx y$.]{\includegraphics[width=\columnwidth]{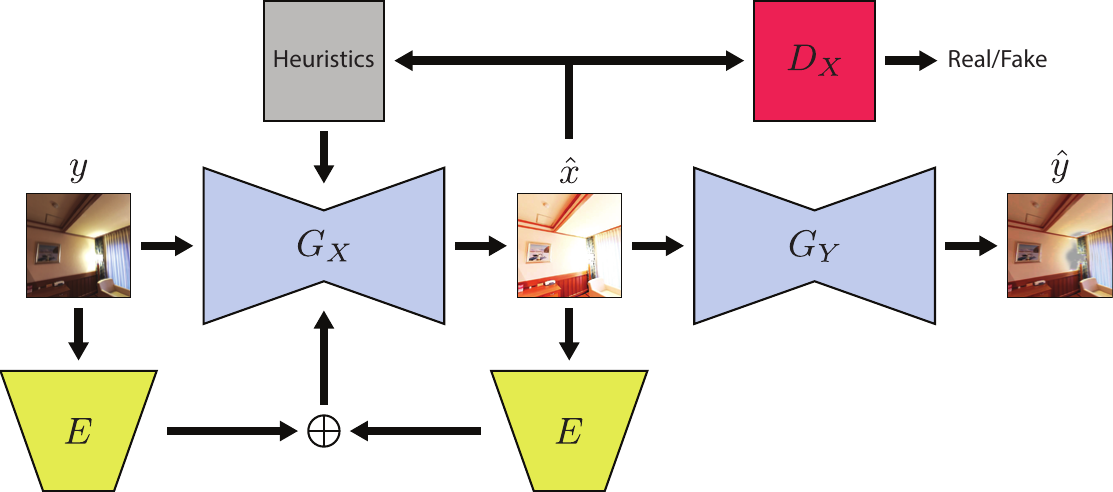}\label{subfig:method_backward}}
\caption{Overview of the proposed CycleHDR architecture (see  \cref{subsec:architecture_modules} and \cref{supsubsec:architecture_modules}) where $x$ and $y$ represent LDR and HDR images, respectively. The method is trained with six objectives: adversarial, cycle consistency, identity, heuristic-based, contrastive, and semantic segmentation (see \cref{subsec:loss_functions} and \cref{supsubsec:loss_functions}).}
\label{fig:method}
\end{figure*}

\noindent\textbf{Semantic and Knowledge-Based Approaches: }
Some methods attempt to use semantic/context information in the image or image formation process for HDR reconstruction.
Wang~\etal~\cite{wang2022kunet} proposed a method that approximates the inverse of the camera pipeline.
Their knowledge-inspired block uses the knowledge of image formation to address three tasks during HDR reconstruction: missing details recovery, adjustment of image parameters, and reduction of image noise.
Goswami~\etal~\cite{goswami2024semantic} proposed a method to recover the clipped intensity values of an LDR/SDR image due to the tone-mapping process in the camera.
The proposed method works in two stages: first it uses a semantic graph-based guidance to help the diffusion process with the in-painting of saturated image parts, and second, the problem is formulated as HDR in-painting from SDR in-painted regions.
Liu~\etal~\cite{liu2022ghost} proposed a vision transformer approach with context-awareness to remove ghosting effects in the output HDR.
The model is a dual-branch architecture to capture both local and global context in the input image that enables the generation process to remove unwanted information in the output HDR and avoid artifacts.

\noindent\textbf{Current Limitations:}
Our analysis reveals that the methods based on single-exposed LDR fail to preserve image characteristics such as the levels of color hue and saturation, contrast, and brightness intensity in HDR.
On the other hand, methods based on multi-exposed LDR produce unwanted visual artifacts such as halo in edges and boundaries, outlier pixel intensities, unwanted repeated pattens, missing details in shadow and highly bright regions, irregular color and texture transitions, and ghosting effects from dynamic scenes.
While most methods produce excellent results, the main limitations of these approaches are the requirement of paired \{LDR;HDR\} datasets for training, and the lack of semantic and contextual knowledge guidance in the reconstruction process that could significantly improve the quality of the HDR output.
Finally, heuristic-based guidance has not been utilized for HDR reconstruction, which might further improve performance.
\cref{tab:related_work} presents a functional comparison between the state-of-the-art and the proposed CycleHDR methods. 

\section{Method}
\label{sec:method}
This section describes CycleHDR, the first semantic and cycle consistency guided self-supervised learning approach for unpaired \{LDR;HDR\} data.
The proposed method is designed to address both the inverse tone-mapping (LDR~$\rightarrow$~HDR) and tone-mapping (HDR~$\rightarrow$~LDR) tasks.

\subsection{Architecture Modules}
\label{subsec:architecture_modules}
We adopt a cycle-consistent adversarial architecture~\cite{zhu2017unpaired} as the basis of our method (see \cref{fig:method}).
The model includes two generators and two discriminators.
The model also takes advantage of CLIP encoders and heuristic-based information to further bridge the gap between semantic and perceptual differences between input and reconstructed images (see \cref{supsubsec:background} for a concise background on the key concepts employed in our method).
Let $X$ and $Y$ be the domains for LDR and HDR images, respectively. 
Furthermore, let $G_{Y}$ denote the generator used in the forward cycle that maps images from LDR to HDR, and let $G_{X}$ be the generator used in the backward cycle that maps images from HDR to LDR.
Moreover, let $D_{X}$ be the discriminator that discriminates between the reconstructed LDR and real LDR images, and let $D_{Y}$ be the discriminator that discriminates between the reconstructed HDR and real HDR images.
The images from the two domains can be represented as $\{x_i\}_{i=1}^N$ where $x_i \in X$ and $\{y_j\}_{j=1}^M$ where $y_j \in Y$.
Similarly, the data distribution of the two domains can be represented as $x \sim p_{\text{data}}(x)$ and $y \sim p_{\text{data}}(y)$.

\noindent\textbf{Generators:}
The generators $G_{Y}$ and $G_{X}$ are based on U-Net architecture~\cite{ronneberger2015u} that includes an encoder and a decoder block with skip connections from each level of the encoder to the decoder.
In our generators, we propose a feedback mechanism~\cite{huang2017densely,yu2015multi} between the encoder and decoder block.
The rationale behind the feedback is to refine the features extracted from the encoder (during the first iteration of the feedback) to guide the decoder for better output image reconstruction over the rest of the iterations.
Hence, the feedback block not only iterates over its own output but also re-runs the decoder situated ahead of it in each iteration while keeping the encoder frozen until the feedback iteration completes.
The feedback is implemented with a ConvLSTM~\cite{huang2017densely,li2019feedback} model and the number of iterations is fixed to $4$ based on an ablation experiment testing $2$, $3$, and $4$ iterations setups (see \cref{supsubsec:architecture_modules} for visual depiction of the proposed generators and implementation details).

\noindent\textbf{Discriminators:}
The discriminators $D_{X}$ and $D_{Y}$ are based on~\cite{li2022uphdr,isola2017image} (see \cref{supsubsec:architecture_modules} for implementation details).

\noindent\textbf{Heuristics:}
The heuristics provide information on whether the generated image includes artifacts and over/underexposed regions for both reconstructed HDR and LDR.
Given that these three factors constitute the nature of the HDR reconstruction, the areas obtained by the heuristics are used in a loss function.
In addition, the heuristics also allow for the generation of a separate saliency map for each of these areas, \ie, artifact, overexposed, and underexposed maps.
Algorithm~\ref{alg:saliency_maps}
illustrates the generation of saliency maps.
The artifact map is multiplied with the input features to the feedback mechanism of the generators $G_{Y}$ and $G_{X}$.
The underexposed map is fed into each of the skip connections (levels $1$ to $3$) in the generator $G_{Y}$ and the overexposed map is multiplied with the input features to the bottleneck layer of this generator.
We use element-wise multiplication as a gating mechanism to emphasize the selected areas in the maps.
The rationale behind the separated saliency map fusion is based on the proposed U-Net architecture where: (a) distorted pixels are mostly refined by the feedback mechanism and later layers; (b) high-intensity lighting details are generally recovered by the middle or later layers, and; (c) low-intensity shadows are reconstructed at the starting and ending layers (see \cref{supsubsec:architecture_modules} for visual depiction of the proposed heuristic-based module and implementation details).

\begin{algorithm}[t]
\caption{Heuristics-Based Saliency Maps}
\label{alg:saliency_maps}
\begin{algorithmic}[1]
\State \textbf{Input:} image\_path, output\_dir
\State \textbf{Output:} Saliency map paths and pixel counts or \texttt{None}
\State Load image and convert to grayscale
\State Generate \textbf{underexposed mask}:
\State \hspace{1em} Mark pixels where intensity $<$ 30
\State Generate \textbf{overexposed mask}:
\State \hspace{1em} Mark pixels where intensity $>$ 240
\State Generate \textbf{artifact mask}:
\State \hspace{1em} Compute Laplacian and Gaussian blur difference
\State \hspace{1em} Mark pixels where edge $>$ 50 and diff $>$ 25
\State Save all masks in \texttt{output\_dir}
\State Return paths and pixel counts for all masks
\end{algorithmic}
\end{algorithm}

\noindent\textbf{Encoders:}
We also introduce a CLIP embeddings encoder~$E$.
Specifically, we use a pre-trained CLIP encoder~\cite{radford2021learning} to extract image embeddings with both local and global semantic context.
For the forward cycle consistency we use $E(x)$ and $E(\hat{y})$.
Then we add the embeddings and feed them back to the bottleneck layer of the $G_{Y}$ decoder.
Similarly, for the backward cycle consistency we add the embeddings from $E(y)$ and $E(\hat{x})$ and feed them back to the bottleneck layer of the $G_{X}$ decoder (see \cref{supsubsec:architecture_modules} for implementation details).

\subsection{Loss Functions}
\label{subsec:loss_functions}
We use six objectives to train the proposed method, including three novel loss functions for HDR reconstruction, namely, heuristic-based, contrastive, and semantic segmentation loss; and two standard CycleGAN loss functions, namely, adversarial loss~\cite{goodfellow2014generative} and cycle consistency loss~\cite{zhu2017unpaired}.
Inspired by~\cite{zhu2017unpaired}, we also use an identity loss~\cite{taigman2016unsupervised}.

The calculations in all loss functions are based on tone-mapped versions of the reconstructed and real HDR images.
This tone-mapping is performed on the basis of the $\mu$-law~\cite{jinno2011mu} and to avoid high-intensity pixel values in HDR images that can distort the loss calculation.
The tone-mapping operator $T$ can be represented as follows:
\begin{equation}
T(y_{j}) = \frac{log(1 + \mu y_{j})}{log(1 + \mu)},
\label{eq:1}
\end{equation}
where the amount of compression $\mu$ is 5000 following~\cite{khan2019fhdr}.

\noindent\textbf{Heuristic-Based Loss:}
Given a reconstructed HDR/LDR image, this loss is based on pixels identified as artifacts and over/underexposed regions by the heuristics in Algorithm~\ref{alg:saliency_maps}.
The loss is defined as:

\begin{equation}
\label{eq:llm_loss}
\mathcal{L}_{\text{heu}} =\delta_1\frac{\hat{y_{\text{af}}}}{\hat{y_{\text{total}}}} + \delta_2\frac{\hat{y_{\text{ox}}}}{\hat{y_{\text{total}}}} + \delta_3\frac{\hat{y_{\text{ux}}}}{\hat{y_{\text{total}}}}, 
\end{equation}
where $\hat{y_{\text{total}}}$ represents the total number of pixels in the reconstructed HDR $\hat{y}$, while $\hat{y_{\text{af}}}$, $\hat{y_{\text{ox}}}$, and $\hat{y_{\text{ux}}}$ denote the total number of pixels in the areas detected by the heuristics as artifacts, overexposed, and underexposed.
The weights on each of the three components are selected after a thorough analysis as well as theoretical considerations.
Our analysis suggests that artifact mitigation is more important than exposure issues for the perceptual quality of the reconstructed images because artifacts make the images unnatural and unpleasing to the human eye, hence $\delta_1$ is set to $3$.
Since humans have greater tolerance towards dark areas than faded bright regions, we give lower weight to the third component, \ie, $\delta_2$ and $\delta_3$ are set to $2$ and $1.5$, respectively (see \cref{supsubsec:loss_functions} for visual depiction of the heuristic-based loss).
This loss is calculated in the backward cycle with $\hat{x}$ but only with the artifact component.
The rationale behind this decision lies in the nature of the output, \ie, LDR images can be overexposed or underexposed.

\noindent\textbf{Contrastive Loss:}
This loss is based on embeddings extracted using a CLIP encoder and ensures semantic information preservation across domains.
Here, we do not directly extract the embeddings from LDR images but instead, use a histogram-equalized version processed using the OpenCV function \texttt{equalizeHist}~\cite{opencv_library}.
Histogram equalization improves the pixel visibility in extreme lighting areas or areas with shadows/darkness of an image by re-adjusting the contrast and saturation levels.
This is done by spreading out the frequent pixel intensity values across $256$ bins.
Equalization often leads to revealing hidden semantic information or non-perceivable objects in an image.
For the image embedding $\bar{x}$ from $E(x)$ and $\bar{y}$ from $E(\hat{y})$ we first define the cosine similarity between them as $\text{sim}(\bar{x},\bar{y})$:

\begin{equation}
\label{eq:six}
\text{sim}(\bar{x},\bar{y}) = \frac{\bar{x} \cdot \bar{y}}{\|\bar{x}\| \|\bar{y}\|},
\end{equation}
where $\cdot$ represents the dot product between the two embeddings and $\|\cdot\|$ represents the norm of the embeddings.

We formulate the contrastive loss for an input batch as positive pairs of images (\ie, LDR and the corresponding reconstructed HDR) and negative pairs of images (\ie, each LDR with the rest of the LDR as well as the rest of the reconstructed HDR in the batch) as 
\begin{equation}
\label{eq:seven}
\begin{split}
 \mathcal{L}_{\text{con}} = -\frac{1}{N} \\
\sum_{i=1}^{N}\log \frac{\exp(\text{sim}(\bar{x}_i,\bar{y}_i)/\tau)}{\sum\limits_{\substack{j=1\\j\neq i}}^{N} (\exp(\text{sim}(\bar{x}_i,\bar{x}_j)/\tau) + \exp(\text{sim}(\bar{x}_i,\bar{y}_j)/\tau))},
\end{split}
\end{equation}
where $N$ is the batch size, $\exp$ represents the exponential function, and $\tau$ represents the temperature parameter which controls the amount of emphasis given in distinguishing between positive and negative pairs (see \cref{subsec:ablation_studies} for analysis of $\tau$).
This loss is calculated in both the forward and backward cycles with the input and output being swapped.
This loss replicates the contrastive learning paradigm of CLIP for \{image;image\} instead of \{image;text\} pairs (see \cref{supsubsec:loss_functions} for visual depiction of the contrastive loss).

\noindent\textbf{Semantic Segmentation Loss:}
This loss is based on segmentation masks.
The Mean Intersection over Union (mIoU) metric measures the amount of overlap between ground truth and predicted segmentation masks.
Similar to the previous loss function, we use histogram-equalized versions of the LDR images processed using the OpenCV function \texttt{equalizeHist}~\cite{opencv_library}.
We choose equalized images instead of original LDR because segmentation of low-light or extremely bright images does not yield good results.
We use the Segment Anything Model~\cite{kirillov2023segment} to generate segmentation classes for the histogram-equalized LDR and reconstructed tone-mapped HDR images (see \cref{supsubsec:loss_functions} for visual depiction of the semantic segmentation loss).
This loss component helps in mitigating differences in boundary and edge pixels between the LDR and HDR images.
We define the per-class IoU metric as $\text{IoU}_c$ and the mean IoU over all segmentation classes as $\text{mIoU}$.
Specifically, the $\text{IoU}_c$ is formulated as:
\begin{equation}
\label{eq:eight}
\text{IoU}_c = \frac{x_c \cap \hat{y_c}}{x_c \cup \hat{y_c}},
\end{equation}
where $x_c$ and $\hat{y_c}$ represent the segmentation for class $c$ in image $x$ and $\hat{y}$, respectively.
$\cap$ represents the overlapping area of predicted and ground truth pixels while $\cup$ represents the total area covered by predicted and ground truth pixels.

The $\text{mIoU}$ is formulated as:
\begin{equation}
\label{eq:nine}
\text{mIoU} = \frac{1}{C} \sum_{c=1}^{C} \text{IoU}_c,
\end{equation}
and we define the semantic segmentation loss as:
\begin{equation}
\label{eq:ten}
\mathcal{L}_{\text{sem}} = 1 - \text{mIoU}.
\end{equation}

This loss is calculated in both forward and backward cycles with the input and output being swapped (see \cref{supsubsec:loss_functions} for visual depiction of the semantic segmentation loss).

\noindent\textbf{Adversarial Loss.}
We apply adversarial loss to both mappings.
First, the mapping from LDR to HDR domain, \ie,  $G_{Y}: X \rightarrow Y$ with the discriminator $D_{Y}$ is expressed as:
\begin{equation}
\label{eq:two}
\begin{split}
\mathcal{L}_{\text{GAN}}(G_{Y},D_{Y},X,Y) = \mathbb{E}_{y \sim p_{\text{data}}(y)}\left[\log D_{Y}(y)\right] + \\
\mathbb{E}_{x \sim p_{\text{data}}(x)}\left[\log(1 - D_{Y}(G_{Y}(x)))\right],
\end{split}
\end{equation}
where $G_{Y}$ generates images that look similar to images from domain $Y$ while $D_{Y}$ distinguishes between generated samples $\hat{y}$ and real samples $y$.
$G_{Y}$ aims to minimize this objective against $D_{Y}$ that, in turn, aims to maximize it, \ie, $\min_{G_{Y}} \max_{D_{Y}} \mathcal{L}_{\text{GAN}}(G_{Y},D_{Y},X,Y)$.
On the other hand, the mapping from HDR to LDR domain, \ie,  $G_{X}: Y \rightarrow X$ with the discriminator $D_{X}$, is expressed as:
\begin{equation}
\label{eq:three}
\begin{split}
\mathcal{L}_{\text{GAN}}(G_{X},D_{X},Y,X) = 
\mathbb{E}_{x \sim p_{\text{data}}(x)}\left[\log D_{X}(x)\right] + \\
\mathbb{E}_{y \sim p_{\text{data}}(y)}\left[\log(1 - D_{X}(G_{X}(y)))\right],
\end{split}
\end{equation}
where $G_{X}$ generates images that look similar to images from domain $X$ while $D_{X}$ distinguishes between generated samples $\hat{x}$ and real samples $x$.
Similar to above, $G_{X}$ aims to minimize this objective against $D_{X}$ that, in turn, aims to maximize it, \ie, $\min_{G_{X}} \max_{D_{X}} \mathcal{L}_{\text{GAN}}(G_{X},D_{X},Y,X)$.

\noindent\textbf{Cycle Consistency Loss:}
The adversarial loss does not guarantee learning without contradiction \ie, the forward $G_{Y}: X \rightarrow Y$ and backward $G_{X}: Y \rightarrow X$ mappings might not be consistent with each other.
Hence, we also incorporate a cycle consistency loss to prevent mutual contradiction of the learned mappings $G_{Y}$ and $G_{X}$.
For each image $x_{i}$ from the LDR domain, the cycle of reconstruction \ie, from LDR to HDR and then back to LDR must result back in the original image $x_{i}$.
Hence, we can define the forward cycle consistency as: $x_{i} \rightarrow G_{Y}(x_{i}) \rightarrow G_{X}(G_{Y}(x_{i})) \approx x_{i}$.
Similarly, the backward cycle consistency can be represented as: $y_{j} \rightarrow G_{X}(y_{j}) \rightarrow G_{Y}(G_{X}(y_{j})) \approx y_{j}$.
The loss is formulated as:
\begin{equation}
\label{eq:four}
\begin{split}
\mathcal{L}_{\text{cyc}}(G_{Y},G_{X}) = 
\mathbb{E}_{x \sim p_{\text{data}}(x)} \left[\|G_{X}(G_{Y}(x)) - x\|_1 \right] + \\
\mathbb{E}_{y \sim p_{\text{data}}(y)} \left[\|G_{Y}(G_{X}(y)) - y\|_1 \right],
\end{split}
\end{equation}
where $\| \cdot \|_1$ is the L1 norm (see \cref{supsubsec:loss_functions} for visual depiction of the cycle consistency loss).

\noindent\textbf{Identity Loss:}
For HDR reconstruction tasks, we also found that adversarial and cycle consistency formulations alone cannot preserve the color and hue information.
This is due to incorrect mapping of color shades from LDR to HDR domains by the generators stemming from the underlying difference in dynamic ranges.
Therefore, we force the generators to replicate an identity mapping by providing target domain images. This loss, denoted by $\mathcal{L}_{\text{id}}(G_{Y},G_{X})$, is expressed as:
\begin{equation}
\label{eq:five}
\mathbb{E}_{y \sim p_{\text{data}}(y)} \left[\|G_{Y}(y) - y\|_1\right] + 
\mathbb{E}_{x \sim p_{\text{data}}(x)} \left[\|G_{X}(x) - x\|_1\right].
\end{equation}

\noindent\textbf{Final Loss:}
The full objective, denoted by $\mathcal{L}_{\text{full}}$, is expressed in \cref{eq:final_loss}, where $\lambda_1$ scales the relative importance of the cycle consistency and identity loss.
$\lambda_1$ is set to $10$ in our experiments and the weight $\lambda_2$ for identity loss is $0.5$ inspired from the setup in the original cycle consistency work~\cite{zhu2017unpaired}.
$\alpha$ and $\beta$ are weights for the contrastive and semantic segmentation loss, respectively.
Both values are set to $2$ (see \cref{subsec:ablation_studies} for analysis of $\alpha$ and $\beta$).
\begin{equation}
\label{eq:final_loss}
\begin{split}
\mathcal{L}_{\text{full}} = 
\mathcal{L}_{\text{GAN}}(G_{Y},D_{Y},X,Y) + \mathcal{L}_{\text{GAN}}(G_{X},D_{X},Y,X) + \\
\lambda_1(\mathcal{L}_{\text{cyc}}(G_{Y},G_{X}) + \lambda_2 \mathcal{L}_{\text{id}}(G_{Y},G_{X})) + \\
\alpha\mathcal{L}_{\text{con}} + \beta\mathcal{L}_{\text{sem}} + \mathcal{L}_{\text{heu}}
\end{split}
\end{equation}

\section{Experiments}
\label{sec:experiments}
\noindent\textbf{Datasets:}
For the primary comparison of our method with the state-of-the-art, we consider the HDRTV~\cite{chen2021new}, NTIRE~\cite{perez2021ntire}, and HDR-Synth \& HDR-Real~\cite{liu2020single} paired datasets.
HDRTV has $1235$ samples for training and $117$ for testing.
The NTIRE dataset consists of approximately $1500$
training, $60$ validation, and $201$ testing samples.
HDR-Synth \& HDR-Real dataset consists of $20537$ samples.
Other paired datasets used in our evaluations are DrTMO~\cite{endo2017deep} ($1043$ samples), Kalantari~\cite{kalantari2017deep} ($89$ samples), HDR-Eye~\cite{korshunov2014crowdsourcing} (46 samples), and LDR-HDR Pair~\cite{jang2020dynamic} ($176$ samples).
We choose these datasets because they consist of a balance between real and synthetic images as well as image and scene diversity.
We used the pre-defined train/test sets of HDRTV and NTIRE.
For HDR-Synth \& HDR-Real, we performed a random $80/20$ split for training and testing in all experiments unless specified otherwise.
For methods working with single-exposed LDR inputs, we use only one LDR from datasets with multi-exposed LDR.
For methods working with multi-exposed LDR inputs, we generate the required exposures using the OpenCV function \texttt{convertScaleAbs}~\cite{opencv_library} for datasets with only single-exposed LDR images.

\noindent\textbf{Metrics:}
We use four metrics to report the results.
High Dynamic Range Visual Differences
Predictor (HDR-VDP-2)~\cite{mantiuk2011hdr} (or Mean Opinion Score Index (Q-Score)) is used for evaluation, replicating the human vision model.
Structural Similarity Index Measure (SSIM)~\cite{wang2009mean,wang2004image,wang2004video} is used to compare block-wise correlations on the basis of structural similarity, luminance, and contrast information.
Peak Signal-to-Noise Ratio (PSNR)~\cite{gupta2011modified} (in dB) is used for a pixel-to-pixel comparison for noise in the signals\footnote{Unless specified otherwise, for HDR evaluation PSNR is computed on tone-mapped original
and reconstructed HDR image pairs.}.
Learned Perceptual Image Patch Similarity (LPIPS)~\cite{zhang2018unreasonable} is used to compare between the high-level features in the images, such as the semantic entities present in the scene, and evaluates our semantic and contextual knowledge-based contributions.
This metric aligns with the perceptual judgment of humans.

\noindent\textbf{Implementation:}
The proposed method is implemented in PyTorch~\cite{paszke2019pytorch} on Ubuntu 20.04.6 LTS workstation with Nvidia Quadro P5000 GPU with 16GB memory, Intel\textregistered~Xeon\textregistered~W-2145 CPU at 3.70GHz with 16 CPU cores, 64GB RAM, and 2.5TB SSD.
We used Adam optimizer~\cite{kingma2014adam} to train the models for $170$ epochs.
Although the batch size of $1$ is expected to work well in cycle consistency models~\cite{zhu2017unpaired}, we set the batch size to $4$ because we are using contrastive loss on positive and negative pairs. 
We used a learning rate of $4 \times 10^{-4}$ for the generators and $2 \times 10^{-4}$ for the discriminators.
The learning rate is kept constant for the first $100$ epochs and subsequently set to decay linearly to $0$.
For all models, we resized the input images to $512 \times 512$.
The HDR images displayed in this manuscript are tone-mapped using Reinhard's operator~\cite{reinhard2023photographic}.

\section{Results}
\label{sec:results}
\subsection{HDR Reconstruction}
\label{subsec:hdr_reconstruction}
For extensive comparisons we selected a combination of methods that utilize different learning paradigms including, HDRCNN~\cite{eilertsen2017hdr}, DrTMO~\cite{endo2017deep}, ExpandNet~\cite{marnerides2018expandnet}, FHDR~\cite{khan2019fhdr}, SingleHDR~\cite{liu2020single}, Two-stage~\cite{a2021two}, KUNet~\cite{wang2022kunet}, HistoHDR-Net~\cite{barua2024histohdr}, HDR-GAN~\cite{niu2021hdr}, ArtHDR-Net~\cite{barua2023arthdr}, Ghost-free HDR~\cite{liu2022ghost}, SelfHDR~\cite{SelfHDR}, UPHDR-GAN~\cite{li2022uphdr}, 
GlowGAN-ITM~\cite{wang2023glowgan}, SingleHDR(W)~\cite{le2023single}, and PSENet~\cite{nguyen2023psenet}.
These state-of-the-art approaches include methods with single-exposed and multi-exposed LDR inputs, direct and indirect (\ie, either directly reconstructing HDR images or generating multi-exposed LDR stacks), GAN-based, Diffusion-based, Transformer-based, unsupervised, weakly-supervised, self-supervised, unpaired learning, and knowledge-conditioned.

\begin{table}[t]
\setlength{\tabcolsep}{1.5pt}
\small
\centering
\caption{HDR reconstruction results. Comparison with supervised (\textcolor{gray}{gray}) and unsupervised, weakly-supervised and self-supervised (\textcolor{black}{black}) learning methods trained and evaluated on the paired datasets HDRTV~\cite{chen2021new}, NTIRE~\cite{perez2021ntire} and HDR-Synth \& HDR-Real~\cite{liu2020single}. \textbf{LP:} Supervised (S), unsupervised (US), weakly-supervised (WS), and self-supervised (SS).}
\label{tab:master_table}
\begin{tabular}{lccccc}
\toprule[0.5mm]
\textbf{Method} & \textbf{LP} & \textbf{PSNR}$\uparrow$ & \textbf{SSIM}$\uparrow$ & \textbf{LPIPS}$\downarrow$ & \textbf{HDR-VDP-2}$\uparrow$ \\
\midrule[0.25mm]
\textcolor{gray}{HDRCNN}~\cite{eilertsen2017hdr} & S & 20.12 & 0.724 & 0.422 & 60.12 \\
\textcolor{gray}{DrTMO}~\cite{endo2017deep} & S & 20.43 & 0.711 & 0.412 & 60.81 \\
\textcolor{gray}{ExpandNet}~\cite{marnerides2018expandnet} &  S & 21.44 & 0.854 & 0.451 & 58.12 \\
\textcolor{gray}{FHDR}~\cite{khan2019fhdr} & S &  20.84 & 0.841 & 0.307 & 62.45 \\
\textcolor{gray}{SingleHDR}~\cite{liu2020single} & S & 21.01 & 0.851 & 0.402 & 64.21 \\
\textcolor{gray}{Two-stage}~\cite{a2021two} & S & 34.29 & 0.856 & 0.287 & 61.78 \\
\textcolor{gray}{HDR-GAN}~\cite{niu2021hdr} & S & 30.11 & 0.912 & 0.411 & 65.01 \\
\textcolor{gray}{KUNet}~\cite{wang2022kunet} & S & 37.21 & 0.974 & 0.051 & 62.14 \\
\textcolor{gray}{Ghost-free}~\cite{liu2022ghost} & S & \textbf{40.32} & 0.966 & 0.043 & 63.02 \\
\textcolor{gray}{ArtHDR-Net}~\cite{barua2023arthdr} & S & 37.12 & 0.961 & 0.321 & 63.43 \\
\textcolor{gray}{HistoHDR-Net}~\cite{barua2024histohdr} & S & 38.21 & 0.968 & 0.301 & 65.15 \\
\midrule[0.125mm]
UPHDR-GAN~\cite{li2022uphdr} & US & 39.98 & 0.965 & 0.044 & 63.54 \\
PSENet~\cite{nguyen2023psenet} & US & 27.35 & 0.856 & 0.274 & 62.89 \\
SingleHDR(W)~\cite{le2023single} & WS & 30.79 & 0.952 & 0.055 & 62.95 \\
GlowGAN-ITM~\cite{wang2023glowgan} & US & 30.19 & 0.901 & 0.064 & 60.05 \\
SelfHDR~\cite{SelfHDR} & SS & 39.51 & 0.972 & 0.037 & 64.77 \\
\midrule[0.25mm]
\textbf{CycleHDR (Ours)} & SS & 40.11 & \textbf{0.979} & \textbf{0.020} & \textbf{68.51} \\
\bottomrule[0.5mm]
\end{tabular}
\end{table}

\begin{table}[t]
\setlength{\tabcolsep}{2pt}
\small
\centering
\caption{HDR reconstruction results on the Kalantari's test set~\cite{kalantari2017deep}. The training datasets are the same as in Table~\ref{tab:master_table}.}
\label{tab:cross-dataset}
\begin{tabular}{lcccc}
\toprule[0.5mm]
\textbf{Method} & \textbf{PSNR}$\uparrow$ & \textbf{SSIM}$\uparrow$ & \textbf{LPIPS}$\downarrow$ & \textbf{HDR-VDP-2}$\uparrow$ \\
\midrule[0.25mm]
Ghost-free~\cite{liu2022ghost} & \textbf{45.03}  & 0.986  & 0.041 & 64.68 \\
SelfHDR~\cite{SelfHDR} & 44.18  & 0.988  & 0.034   & 64.86 \\
\midrule[0.25mm]
\textbf{CycleHDR (Ours)} & 44.76 & \textbf{0.989} & \textbf{0.020}  & \textbf{68.85} \\
\bottomrule[0.5mm]
\end{tabular}
\end{table}

\begin{figure}[t]
\centering
\includegraphics[width=\columnwidth]{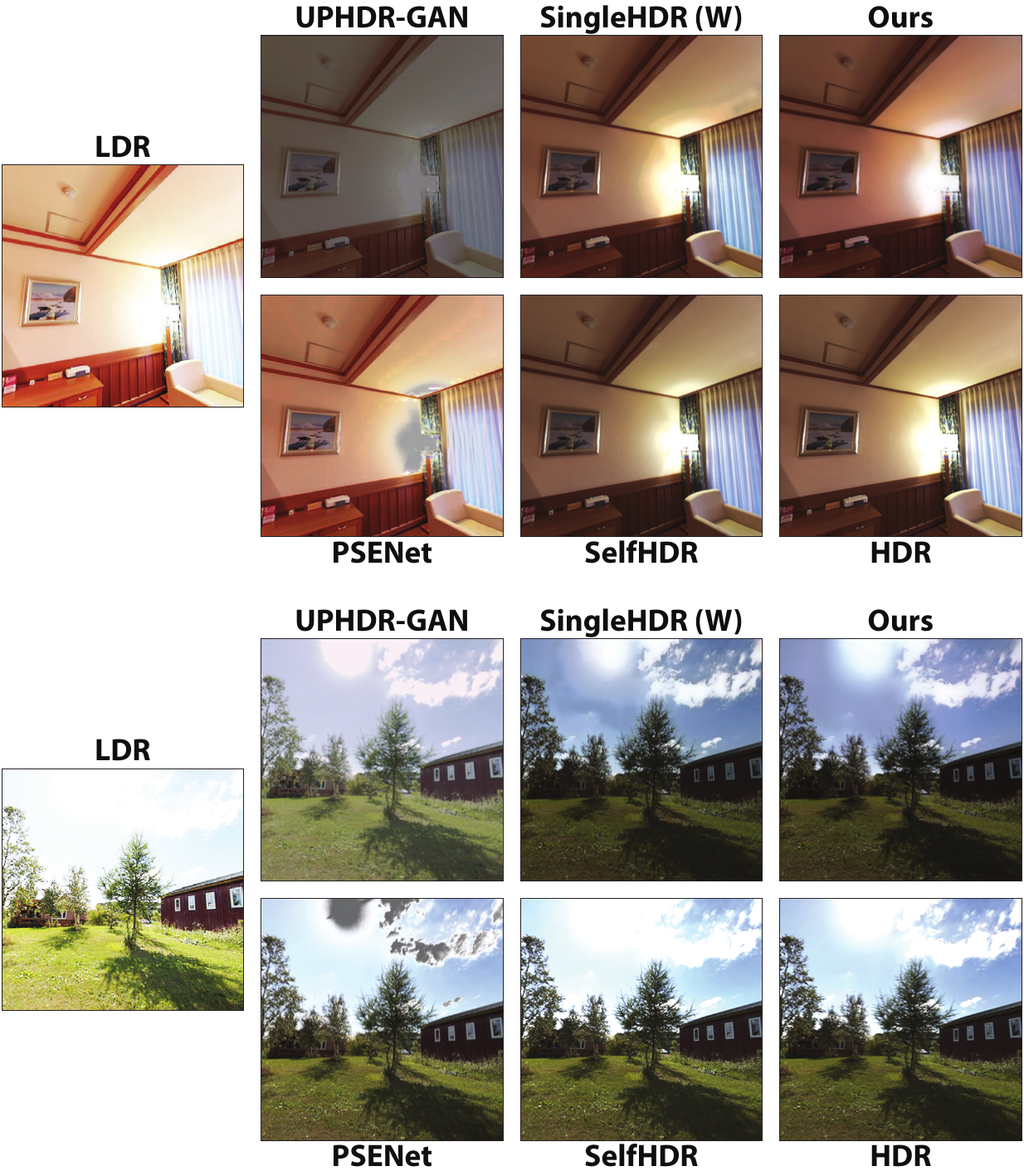}
\caption{Examples of HDR images reconstructed with our method and recent state-of-the-art methods.\label{fig:results1}}
\end{figure}

\noindent\textbf{Paired Datasets:}
\cref{tab:master_table} summarizes the results of evaluation on HDRTV~\cite{chen2021new}, NTIRE~\cite{perez2021ntire}, HDR-Synth \& HDR-Real~\cite{liu2020single} paired datasets, the most widely used datasets by the state-of-the-art.
Fig.~\ref{fig:results1} illustrates the quality of the HDR images reconstructed with our method (see \cref{supsec:results} for an extensive qualitative evaluation).
In the top part of the table, we compare our method with $11$ supervised learning approaches.
The proposed method outperforms those approaches in terms of SSIM, LPIPS, and HDR-VDP-2.
In terms of PSNR, ours is the second best after Ghost-free~\cite{liu2022ghost}.
In the bottom part of the table, we compare our method with $5$ unsupervised, weakly-supervised, and self-supervised learning methods.
The proposed method outperforms those approaches on all evaluation metrics.
Although the improvements are small for PSNR and SSIM, LPIPS and HDR-VDP-2 show considerable improvements, which is important to note, as perceptual, semantic, and high dynamic qualities are captured by these two metrics only.
It is noteworthy that we used paired datasets in order to compare our method with the state-of-the-art, which predominantly requires paired samples for training.
However, in all experiments, our method is trained in a self-supervised fashion with unpaired samples (see~\cref{sec:method}).
To further evaluate the performance of our method, we report the results of a cross-dataset (unseen samples) experiment using Kalantari's test set~\cite{kalantari2017deep}.
Tab.~\ref{tab:cross-dataset} records the results, which suggest that our method outperforms two state-of-the-art methods, Ghost-free~\cite{liu2022ghost} and SelfHDR~\cite{SelfHDR}, on all metrics except PSNR.

\begin{table}[t]
\setlength{\tabcolsep}{2pt}
\small
\centering
\caption{HDR reconstruction results. Comparison with unsupervised learning methods trained on mixed datasets.}
\label{tab:unpaired_test}
\begin{tabular}{lcccc}
\toprule[0.5mm]
\textbf{Method} & \textbf{PSNR}$\uparrow$ & \textbf{SSIM}$\uparrow$ & \textbf{LPIPS}$\downarrow$ & \textbf{HDR-VDP-2}$\uparrow$ \\
\midrule[0.25mm]
UPHDR-GAN~\cite{li2022uphdr} & 42.51 & 0.980 & 0.043 & 66.13 \\
GlowGAN-ITM~\cite{wang2023glowgan} & 33.45 & 0.910 & 0.061 & 62.13 \\
\midrule[0.25mm]
\textbf{CycleHDR (Ours)} & \textbf{43.96} & \textbf{0.989} & \textbf{0.020}  & \textbf{69.32} \\
\bottomrule[0.5mm]
\end{tabular}
\end{table}

\begin{table}[ht]
\setlength{\tabcolsep}{2pt}
\small
\centering
\caption{HDR reconstruction results. Comparison with unsupervised learning methods trained on unpaired datasets.}
\label{tab:cross_dataset_test}
\begin{tabular}{lcccc}
\toprule[0.5mm]
\textbf{Method} & \textbf{PSNR}$\uparrow$ & \textbf{SSIM}$\uparrow$ & \textbf{LPIPS}$\downarrow$ & \textbf{HDR-VDP-2}$\uparrow$ \\
\midrule[0.25mm]
UPHDR-GAN~\cite{li2022uphdr} & 37.51 & 0.941 & 0.101 & 61.63 \\
GlowGAN-ITM~\cite{wang2023glowgan} & 32.15 & 0.923 & 0.137 & 60.11 \\
\midrule[0.25mm]
\textbf{CycleHDR (Ours)} & \textbf{39.32} & \textbf{0.952} & \textbf{0.082}  & \textbf{64.32} \\
\bottomrule[0.5mm]
\end{tabular}
\end{table}

\begin{table}[t]
\setlength{\tabcolsep}{2pt}
\small
\centering
\caption{LDR reconstruction results. Comparison with the state-of-the-art tone-mapping operators and methods.}
\label{tab:ldr_test}
\begin{tabular}{lccc}
\toprule[0.5mm]
\textbf{Method} & \textbf{PSNR}$\uparrow$ & \textbf{SSIM}$\uparrow$ & \textbf{LPIPS}$\downarrow$ \\
\midrule[0.25mm]
$\mu$-law operator~\cite{jinno2011mu} & 28.68 & 0.891 & 0.311 \\
Reinhard's operator~\cite{reinhard2023photographic} & 30.12 & 0.903 & 0.342 \\
Photomatix~\cite{photomatix} & 32.11 & 0.922 & 0.136 \\
DeepTMO~\cite{rana2019deep} & \textbf{32.23} & 0.924 & 0.132 \\
Unpaired-TM~\cite{vinker2021unpaired} & 31.71 & 0.925 & 0.127 \\
Zero-shot-TM~\cite{zhu2024zero} & 31.75 & 0.936 & 0.107 \\
GlowGAN-prior~\cite{wang2023glowgan} & 32.03 & 0.930 & 0.108 \\
\midrule[0.25mm]
\textbf{CycleHDR (Ours)} & 32.11 & \textbf{0.942} & \textbf{0.091} \\
\bottomrule[0.5mm]
\end{tabular}
\end{table}

\noindent\textbf{Mixed Datasets:}
To further demonstrate the strength of our approach, we extended the train set used in the previous experiment (paired datasets) with: 1) HDR images from DrTMO~\cite{endo2017deep}, Kalantari~\cite{kalantari2017deep}, HDR-Eye~\cite{korshunov2014crowdsourcing}, LDR-HDR Pair~\cite{jang2020dynamic}, and GTA-HDR~\cite{barua2024gta} ($1500$ random samples), \ie, additional $2854$ HDR images; and 2) LDR images from GTA-HDR ($2000$ random samples) and $50\%$ of the LDR images in DrTMO, Kalantari, HDR-Eye, and LDR-HDR Pair, \ie, additional $2677$ LDR images.
There is no overlap between LDR and HDR in this additional data, \ie, it is an unpaired dataset.
The final dataset used for training in this experiment has an approximately 87\% overlap between LDR and HDR images.
We kept the test set identical to the previous experiment, reported in \cref{tab:master_table}, in order to compare with the state-of-the-art methods based on paired data.
Given that the only other methods that support this type of unpaired train data are UPHDR-GAN~\cite{li2022uphdr} and GlowGAN-ITM~\cite{wang2023glowgan}, we perform a direct comparison with these approaches in \cref{tab:unpaired_test}.
The results demonstrate that our method outperforms all unpaired and paired data methods, setting new state-of-the-art on all metrics (\cref{supsubsec:hdr_reconstruction} for more mixed data experiments).

\noindent\textbf{Unpaired Datasets:\label{subsec:cross_data_reconstruction}}
We performed a fully unpaired data experiment where LDR and HDR for training are sourced from different datasets.
We kept the test set identical to the previous experiments.
LDR images are sourced from HDRTV~\cite{chen2021new} and NTIRE~\cite{perez2021ntire}, while the HDR images are taken from HDR-Synth \& HDR-Real~\cite{liu2020single}.
The direct comparison with UPHDR-GAN~\cite{li2022uphdr} and GlowGAN-ITM~\cite{wang2023glowgan}, which have unpaired training capabilities, is reported in \cref{tab:cross_dataset_test}.
The results demonstrate that our method outperforms both unpaired data methods on all metrics.

\subsection{LDR Reconstruction}
\label{subsec:ldr_reconstruction}
We report results for LDR reconstruction in \cref{tab:ldr_test} and compare with the state-of-the-art tone-mapping operators $\mu$-law~\cite{jinno2011mu}, Reinhard's~\cite{reinhard2023photographic}, and Photomatix~\cite{photomatix}.
We also consider the HDR-to-LDR prior from GlowGAN~\cite{wang2023glowgan} which is used to re-project reconstructed HDR to LDR space for inverse learning. Additionally we include recent HDR tone-mapping methods such as Unpaired-TM~\cite{vinker2021unpaired}, DeepTMO~\cite{rana2019deep} and Zero-shot-TM~\cite{zhu2024zero} for comparison. The test data is identical to the one used in the HDR reconstruction experiments.
The results demonstrate that our method outperforms all considered methods in terms of SSIM and LPIPS.
In terms of PSNR, ours is the second best after DeepTMO (see \cref{supsubsec:ldr_reconstruction} for more LDR reconstruction experiments).

\subsection{Ablation Results\label{subsec:ablation_studies}}
\begin{figure}[t]
\centering
\includegraphics[width=\columnwidth]{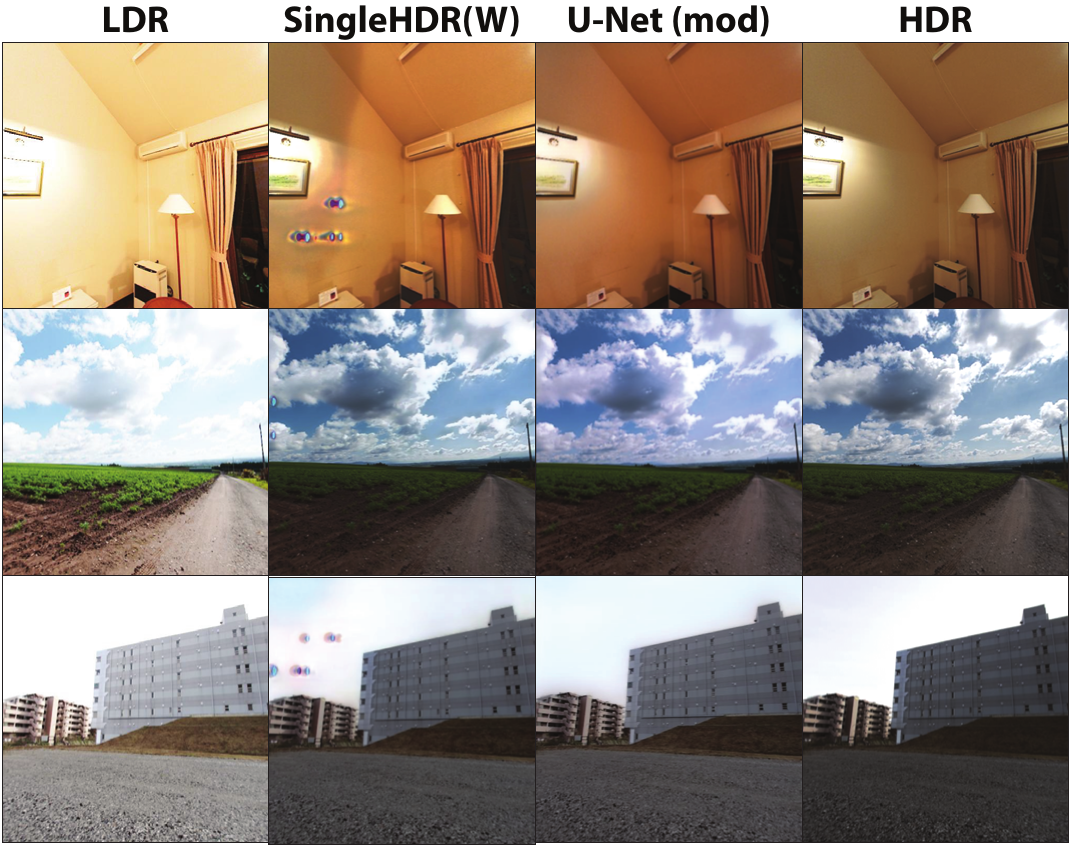}
\caption{Examples of HDR images using the SingleHDR(W)~\cite{le2023single} U-Net with and without our feedback mechanism on images from the DrTMO~\cite{endo2017deep}.\label{fig:ablation_u_net}}
\end{figure}

\noindent\textbf{Generators:}
We propose a novel generator that modifies the U-Net architecture by introducing an artifact-aware feedback mechanism (see \cref{subsec:architecture_modules}).
The feedback helps in avoiding artifacts in image-to-image translation tasks, where a huge volume of training data might make the process of hallucinating details difficult.
To test this hypothesis, we first replace the original U-Net used in the SingleHDR(W)~\cite{le2023single} method with our U-Net generator.
We see that this leads to improvements in all but one metric (\ie, PSNR), as shown in \cref{tab:sub_table2}.
The table also demonstrates the improvement in our model when we use the feedback U-Net instead of the original U-Net of SingleHDR(W). 
\cref{fig:ablation_u_net} illustrates the improvement in SingleHDR(W)~\cite{le2023single} when we use the proposed feedback U-Net instead of the original U-Net of SingleHDR(W).
The original U-Net produces many artifacts in the output HDR images whereas our modified version with feedback reconstructs artifact-free HDR images.

\begin{table}[t]
\setlength{\tabcolsep}{2pt}
\small
\centering
\caption{Quantitative evaluation of our feedback U-Net-based generators for HDR reconstruction. For evaluation, we use the DrTMO~\cite{endo2017deep} dataset. The models marked with ``(mod)" use feedback U-Net-based generators and the rest use the same U-Net generator as SingleHDR(W).\label{tab:sub_table2}}
\begin{tabular}{lcccc}
\toprule[0.5mm]
\textbf{Method} & \textbf{PSNR}$\uparrow$ & \textbf{SSIM}$\uparrow$ & \textbf{LPIPS}$\downarrow$ & \textbf{HDR-VDP-2}$\uparrow$ \\
\midrule[0.25mm]
SingleHDR(W)~\cite{le2023single} & 30.79 & 0.952 & 0.055 & 62.95 \\
SingleHDR(W) (mod) & 30.03 & 0.961 & 0.036 & 63.91 \\
\midrule[0.25mm]
\textbf{CycleHDR} & 35.11 & 0.969 & 0.029 & 67.88 \\
\textbf{CycleHDR (mod)} & \textbf{43.01} & \textbf{0.988} & \textbf{0.023} & \textbf{68.21} \\
\bottomrule[0.5mm]
\end{tabular}
\end{table}

\noindent\textbf{Architecture:}
Our method uses the cycle consistency concept of CycleGAN and introduces many new components to address the LDR~$\leftrightarrow$~HDR reconstruction tasks.
The most significant components are the heuristics module, artifact-aware feedback and exposure-aware skip connections-based U-Net generators, as well as the CLIP embedding encoders (see \cref{subsec:architecture_modules}).
We perform a thorough ablation analysis on each of the components to highlight their contributions.
Tab.~\ref{tab:arch_ablation} summarizes the results where we add each component to the architecture as we go down the rows.
The first row lists the results from the original CycleGAN method.
The second row is the CycleGAN utilizing the U-Net generator from SingleHDR(W).
The third row is the CycleGAN utilizing our feedback U-Net generator.
This results in a significant improvement in SSIM, LPIPS, and HDR-VDP-2.
The fourth row shows the results when we add the heuristics module to the architecture, which shows some improvement in PSNR, LPIPS, and HDR-VDP-2.
The fifth row records the results when we remove the heuristics module and add the CLIP embedding encoders to the architecture.
This results in an improvement in PSNR, SSIM, and HDR-VDP-2.
The sixth row depicts improvement in almost all the metrics as we add both CLIP encoders and the heuristics module to the architecture.
The seventh, eighth, and ninth rows depict the improvements after adding contrastive, semantic segmentation, and heuristic-based objectives (see \cref{subsec:loss_functions}) in the architecture, respectively.
The tenth row shows the improvements when we add only the contrastive and semantic segmentation objectives in the architecture.
The eleventh row depicts the results with all components in our architecture, but with the original U-Net from SingleHDR(W).
The last row is the full architecture with all components, which achieves the best performance on all metrics.

\begin{table}[t]
\setlength{\tabcolsep}{1pt}
\small
\centering
\caption{Quantitative evaluation of different components. For evaluation we use the HDRTV~\cite{chen2021new}, NTIRE~\cite{perez2021ntire} and HDR-Synth \& HDR-Real~\cite{liu2020single} datasets. \textbf{U\textsuperscript{o}:} The U-Net generator used in SingleHDR(W)~\cite{le2023single}, \textbf{U\textsuperscript{f}:} The proposed U-Net generator, \textbf{E:} CLIP embedding encoder, and \textbf{Heu (Heuristic):} Heuristics module.}
\label{tab:arch_ablation}
\begin{tabular}{lcccc}
\toprule[0.5mm]
\textbf{Variant} & \textbf{PSNR}$\uparrow$ & \textbf{SSIM}$\uparrow$ & \textbf{LPIPS}$\downarrow$ & \textbf{HDR-VDP-2}$\uparrow$ \\
\midrule[0.25mm]
CycleGAN~\cite{zhu2017unpaired} & 21.34 & 0.849 & 0.351 & 61.55 \\
U\textsuperscript{o} & 25.13 & 0.851 & 0.341 & 61.79 \\
U\textsuperscript{f} & 24.07 & 0.879 & 0.126 & 63.16 \\
U\textsuperscript{f}+Heu & 26.12 & 0.880 & 0.121 & 64.32 \\
U\textsuperscript{f}+E & 29.19 & 0.885 & 0.122 & 64.21 \\
U\textsuperscript{f}+E+Heu & 30.15 & 0.887 & 0.101 & 64.41 \\
U\textsuperscript{f}+E+Heu+$\mathcal{L}_{\text{con}}$ & 36.45 & 0.911 & 0.047 & 66.56 \\
U\textsuperscript{f}+E+Heu+$\mathcal{L}_{\text{sem}}$ & 34.21 & 0.897 & 0.094 & 66.12 \\
U\textsuperscript{f}+E+Heu+$\mathcal{L}_{\text{Heu}}$ & 36.01 & 0.907 & 0.046 & 66.72 \\
U\textsuperscript{f}+E+Heu+$\mathcal{L}_{\text{con}}$+$\mathcal{L}_{\text{sem}}$ & 40.01 & 0.979 & 0.021 & 68.41\\
U\textsuperscript{o}+E+Heu+$\mathcal{L}_{\text{con}}$+$\mathcal{L}_{\text{sem}}$+$\mathcal{L}_{\text{Heu}}$ & 35.15 & 0.953 & 0.042 & 67.18 \\
\midrule[0.25mm]
U\textsuperscript{f}+E+Heu+$\mathcal{L}_{\text{con}}$+$\mathcal{L}_{\text{sem}}$+$\mathcal{L}_{\text{Heu}}$ & \textbf{40.11} & \textbf{0.979} & \textbf{0.020} & \textbf{68.51} \\
\bottomrule[0.5mm]
\end{tabular}
\end{table}

\begin{table}[t]
\setlength{\tabcolsep}{2pt}
\small
\centering
\caption{Loss functions ablation results with the HDRTV~\cite{chen2021new}, NTIRE~\cite{perez2021ntire} and HDR-Synth \& HDR-Real~\cite{liu2020single} datasets.}
\label{tab:loss}
\begin{tabular}{lcccc}
\toprule[0.5mm]
\textbf{Loss} & \textbf{PSNR}$\uparrow$ & \textbf{SSIM}$\uparrow$ & \textbf{LPIPS}$\downarrow$ & \textbf{HDR-VDP-2}$\uparrow$ \\
\midrule[0.25mm]
$\mathcal{L}_{\text{GAN}}$+$\mathcal{L}_{\text{cyc}}$ & 24.55 & 0.878 & 0.125 & 62.11 \\
+$\mathcal{L}_{\text{id}}$ & 30.15 & 0.887 & 0.101 & 64.41 \\
+$\mathcal{L}_{\text{id}}$+$\mathcal{L}_{\text{con}}$ & 36.45 & 0.911 & 0.047 & 66.56 \\
+$\mathcal{L}_{\text{id}}$+$\mathcal{L}_{\text{sem}}$ & 34.21 & 0.897 & 0.094 & 66.12 \\
+$\mathcal{L}_{\text{id}}$+$\mathcal{L}_{\text{Heu}}$ & 36.01 & 0.907 & 0.046 & 66.72 \\
+$\mathcal{L}_{\text{id}}$+$\mathcal{L}_{\text{con}}$+$\mathcal{L}_{\text{sem}}$ & 40.01 & 0.979 & 0.021 & 68.41 \\
+$\mathcal{L}_{\text{id}}$+$\mathcal{L}_{\text{con}}$+$\mathcal{L}_{\text{sem}}$+$\mathcal{L}_{\text{Heu}}$ & \textbf{40.11} & \textbf{0.979} & \textbf{0.020} & \textbf{68.51} \\
\bottomrule[0.5mm]
\end{tabular}
\end{table}

\begin{figure}[t]
\centering
\includegraphics[width=\columnwidth]{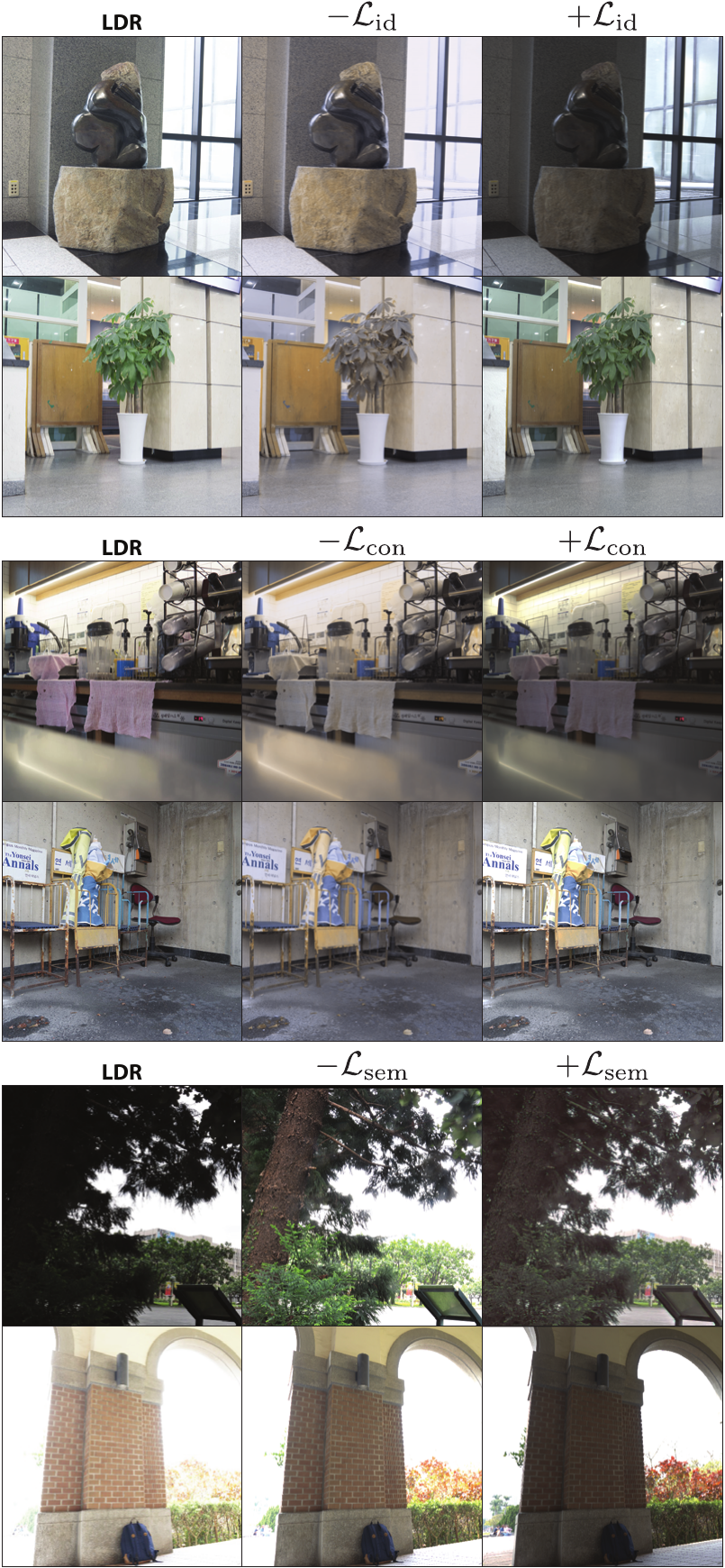}
\caption{Ablation results for $\mathcal{L}_{\text{id}}$, $\mathcal{L}_{\text{con}}$, and $\mathcal{L}_{\text{sem}}$ with images from the HDR-Synth \& HDR-Real~\cite{liu2020single} and LDR-HDR pair datasets~\cite{jang2020dynamic}.}
\label{fig:ablation_id_loss}
\end{figure}

\noindent\textbf{Loss Functions:}
We study the loss functions and their contribution towards the overall results, keeping the architectural components constant.
\cref{tab:loss} summarizes the results.
The first row is the results from the standard adversarial and cycle consistency loss.
In the second row, we add the identity loss, which improves all metrics.
In the third, fourth, and fifth rows, we add the contrastive, semantic segmentation, and heuristic-based objectives.
This results in an improvement in all metrics.
The sixth row shows the results when we add contrastive and semantic losses together, which gives significant improvement in the metrics.
Finally, adding all loss functions (seventh row) achieves the best results on all metrics.
\cref{fig:ablation_id_loss} visually validates the contribution of the identity loss $\mathcal{L}_{\text{id}}$, illustrating that it helps in recovering the hue and shades of the original image in the reconstruction.
It further demonstrates similar results for the contrastive and semantic segmentation objectives.
The contrastive loss $\mathcal{L}_{\text{con}}$ can recover high-level attributes such as color and texture information more accurately compared to our method without this loss component.
We also observe that the semantic segmentation loss $\mathcal{L}_{\text{sem}}$ recovers the low-level attributes such as object boundaries and edges more vividly than our method without this loss component.
These high/low-level attributes are of vital importance to downstream applications such as object recognition.

We also perform separate tests on contrastive and semantic segmentation loss to examine the contribution of histogram-equalized LDR compared to original LDR images.
We summarize these results in \cref{tab:sub_table5} and \cref{tab:sub_table4}.
In both cases, the loss calculated between the histogram-equalized versions of the LDR (instead of the original LDR) and the reconstructed HDR is better.
This is due to the fact that histogram-equalized LDR can reveal semantic information in extremely over/underexposed areas of the original LDR images, which in turn improves the extracted CLIP embeddings for $\mathcal{L}_{\text{con}}$ and segmentation maps for $\mathcal{L}_{\text{sem}}$.

\begin{figure*}[t]
\captionsetup[subfigure]{justification=centering}
\centering
\subfloat[]{\includegraphics[width=0.3\textwidth]{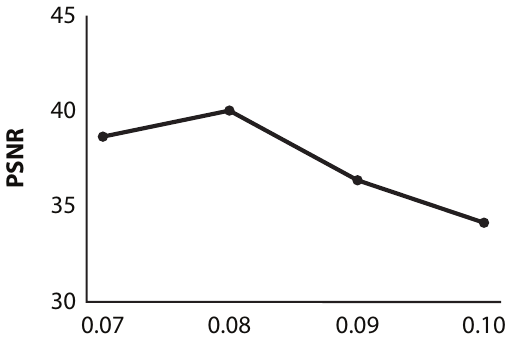}}
\hfill
\subfloat[]{\includegraphics[width=0.3\textwidth]{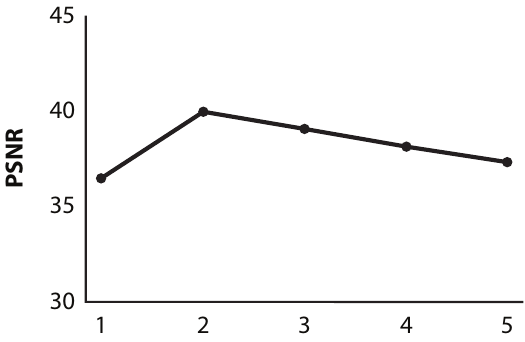}}
\hfill
\subfloat[]{\includegraphics[width=0.3\textwidth]{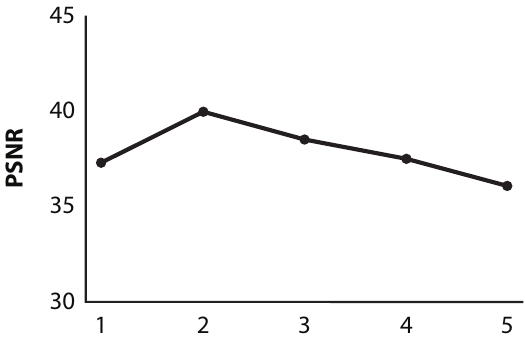}}
\caption{Ablation results for the temperature parameter $\tau$ in (a), and loss component weights $\alpha$ in (b) and $\beta$ in (c).}
\label{fig:ablation}
\end{figure*}

Finally, we experiment with the weights for the loss components with respect to PSNR.
We perform tests on the temperature parameter $\tau$ of the contrastive loss and the weight parameters $\alpha$ and $\beta$ for the loss components $\mathcal{L}_{\text{con}}$ and $\mathcal{L}_{\text{sem}}$.
\cref{fig:ablation} shows the results.
The temperature parameter $\tau$ gives the optimal PSNR with $\tau=0.08$ while the result degrades if we increase the value further.
If we decrease the value, it gives a sub-optimal PSNR at $\tau=0.07$, so we set the value to $\tau=0.08$ in all experiments.
When $\alpha=2$ and $\beta=2$, the optimal PSNR value is attained. 
The weights for the heuristic-based loss are set to $3$, $2$, and $1.5$ for the artifact, overexposed, and underexposed components, respectively.
We conducted ablation experiments on these weights, considering values from $1$ to $5$, and found that the selected combination of weights yields the best PSNR.

\begin{table}[t]
\setlength{\tabcolsep}{2pt}
\small
\centering
\caption{Semantic segmentation loss ablation results with the HDRTV~\cite{chen2021new}, NTIRE~\cite{perez2021ntire} and HDR-Synth \& HDR-Real~\cite{liu2020single} datasets.}
\label{tab:sub_table4}
\begin{tabular}{lcccc}
\toprule[0.5mm]
\textbf{Variant} & \textbf{PSNR}$\uparrow$ & \textbf{SSIM}$\uparrow$ & \textbf{LPIPS}$\downarrow$ & \textbf{HDR-VDP-2}$\uparrow$ \\
\midrule[0.25mm]
LDR & 38.89 & 0.956 & 0.033 & 66.92 \\
LDR\textsuperscript{Hist} & \textbf{40.11} & \textbf{0.979} & \textbf{0.020} & \textbf{68.51} \\
\bottomrule[0.5mm]
\end{tabular}
\end{table}

\begin{table}[t]
\setlength{\tabcolsep}{2pt}
\small
\centering
\caption{Contrastive loss ablation results with the HDRTV~\cite{chen2021new}, NTIRE~\cite{perez2021ntire} and HDR-Synth \& HDR-Real~\cite{liu2020single} datasets.}
\label{tab:sub_table5}
\begin{tabular}{lcccc}
\toprule[0.5mm]
\textbf{Variant} & \textbf{PSNR}$\uparrow$ & \textbf{SSIM}$\uparrow$ & \textbf{LPIPS}$\downarrow$ & \textbf{HDR-VDP-2}$\uparrow$ \\
\midrule[0.25mm]
LDR & 39.12 & 0.966 & 0.030 & 67.11 \\
LDR\textsuperscript{Hist} & \textbf{40.11} & \textbf{0.979} & \textbf{0.020} & \textbf{68.51} \\
\bottomrule[0.5mm]
\end{tabular}
\end{table}

\section{Limitations and Future Work}
\label{sec:future}
Our method is designed for LDR~$\leftrightarrow$~HDR bi-directional reconstruction of images and can be further extended to videos with some modifications in the generator models (\eg, using special ViT~\cite{han2022survey} or Timesformer~\cite{bertasius2021space}).
Further research can also explore multi-task architectures along with dynamic range conversion such as deblurring, denoising, super-resolution, and demosaicing tasks.
The pre-processing step in our loss calculations for contrastive and semantic segmentation loss employs histogram equalization, which can be further extended with methods like Gamma correction~\cite{li2022gamma} or histogram matching~\cite{coltuc2006exact}.
Another direction can be the integration of no-reference quality assessment~\cite{banterle2020nor} into the framework which will enable parallel training and prediction of the reconstructed LDR/HDR along with its quality score.
Instead of the heuristics formulation to make the generators artifact- and exposure-aware, more advanced formulations such as explainability of the generation process and object's material-based specular reflections can also be introduced in the future.
Furthermore, the differentiable ray tracing technology~\cite{glassner1989introduction} deployed to generate realistic HDR images/videos in video games~\cite{barua2024gta} to simulate the physics behind light, specular reflections, and shadows can also be explored using deep learning concepts such as Nvdiffrast~\cite{laine2020modular} and NeRF~\cite{mildenhall2022nerf}.

\section{Conclusion\label{sec:conclusion}}
This paper proposes CycleHDR, the first semantic and cycle consistency guided self-supervised learning approach for unpaired \{LDR;HDR\} data which addresses both the inverse tone-mapping (\ie, LDR~$\rightarrow$~HDR) and tone-mapping (\ie, HDR~$\rightarrow$~LDR) tasks.
The proposed method utilizes novel generators based on modified U-Net architecture incorporating ConvLSTM-based artifact-aware feedback mechanism and exposure-aware skip connections to mitigate visual artifacts, CLIP embedding encoder for contrastive learning to minimize the semantic difference between LDR and reconstructed HDR pairs, and a novel loss function based on the Mean Intersection over Union metric to further ensure semantic consistency between the LDR and reconstructed HDR.
It also utilizes heuristics to define a loss and artifact- and exposure-aware saliency maps to bring more realism and naturalness in the reconstructed HDR images.
The thorough experimental validation demonstrates the contributions of the proposed method that achieves state-of-the-art results across several benchmark datasets and reconstructs high-quality HDR and LDR images.

\bibliographystyle{IEEEtran}
\bibliography{bibliography}

\clearpage

 
\vspace{11pt}


\vspace{11pt}


\title{A Cycle Ride to HDR: Semantics Aware Self-Supervised Framework for Unpaired LDR-to-HDR Image Reconstruction}

\maketitle










\setcounter{page}{1}

\setcounter{section}{0}
\setcounter{figure}{0}
\setcounter{table}{0}
\renewcommand*{\thesection}{S\arabic{section}}
\renewcommand*{\thefigure}{S\arabic{figure}}
\renewcommand*{\thetable}{S\arabic{table}}

\section{Method}
\label{supsec:method}
This section complements Sec.~III in the main paper.
It provides a concise background on the key concepts employed in our method, \ie, cycle consistency~\cite{zhu2017unpaired}, contrastive language-image pretraining~\cite{radford2021learning,lisupervision,yee2024clipswap}, heuristics~\cite{opencv_library}, and image semantic segmentation~\cite{kirillov2023segment}.
The section also offers implementation details for the architecture modules.

\subsection{Background}
\label{supsubsec:background}
\begin{figure}[t]
\centering
\includegraphics[width=\linewidth]{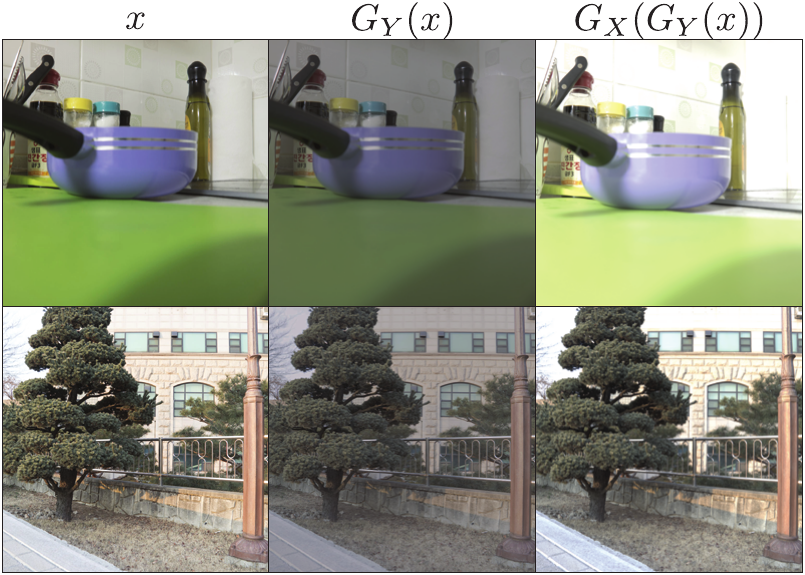}
\caption{$G_{Y}$ translates LDR to HDR and $G_{X}$ HDR to LDR. The LDR images are from the LDR-HDR pair~\cite{jang2020dynamic} dataset.}
\label{fig:cycle_consistency}
\end{figure}

\noindent\textbf{Cycle Consistency.}
The concept of cycle consistency in image-to-image~\cite{zhu2017unpaired} and video-to-video~\cite{chen2019mocycle} translation tasks is a powerful and elegant design for unpaired data.
Previously, cycle consistency has been applied to paired LDR $\leftrightarrow$ HDR translation~\cite{yan2021towards}, where the authors used $3$ LDR images with different exposures as input to reconstruct an HDR.
The reconstructed HDR is then used by $3$ different generators to reconstruct the original $3$ multi-exposed LDR images.
However, the focus of our work is on the more general task of unpaired LDR $\leftrightarrow$ HDR translation.
Given the images $\{x_i\}_{i=1}^N$ where $x_i \in X$ and $\{y_j\}_{j=1}^M$ where $y_j \in Y$, the goal is to design two generators, \ie, $G_Y$ and $G_X$ such that, $G_Y(x)$ $\rightarrow$ $y$ and $G_X(y)$ $\rightarrow$ $x$.
\cref{fig:cycle_consistency} depicts this concept for the LDR-to-HDR translation task.

\noindent\textbf{Contrastive Language-Image Pretraining.}
CLIP~\cite{radford2021learning} is a method that finds the closeness between visual and textual data in the embedding space.
The method is trained on image-text pairs to associate images with their descriptions.
The CLIP loss maximizes the cosine similarity between correct (positive) pairs and minimize the similarity between incorrect (negative) pairs.
We extract the CLIP embeddings from both the LDR and reconstructed HDR image pairs and maximize their cosine similarity in the embedding space for preservation of semantic information across domains.
We also combine the CLIP embeddings extracted from the LDR and reconstructed HDR to guide the decoder of the generators in the next step of the reconstruction.

\noindent\textbf{Semantic Segmentation.}
Semantic segmentation~\cite{kirillov2023segment,ravi2024sam,li2023semantic} is a vision problem where each pixel of an image or video frame is segmented into meaningful sub-parts such as objects, roads, water bodies, buildings, humans, and animals.
The Segment Anything Model~\cite{kirillov2023segment} is a state-of-the-art method for semantic segmentation of images.
Segmented objects in a scene not only inform the number and location of semantically meaningful concepts, but also provide a holistic semantic description of the scene.
Therefore, we can use that information to improve image reconstruction.
We introduce a loss on the basis of the Mean Intersection over Union metric to preserve the image integrity (based on the segmented objects) between the two domain $X$ and $Y$.

\begin{figure*}[t]
\centering
\includegraphics[width=\linewidth]{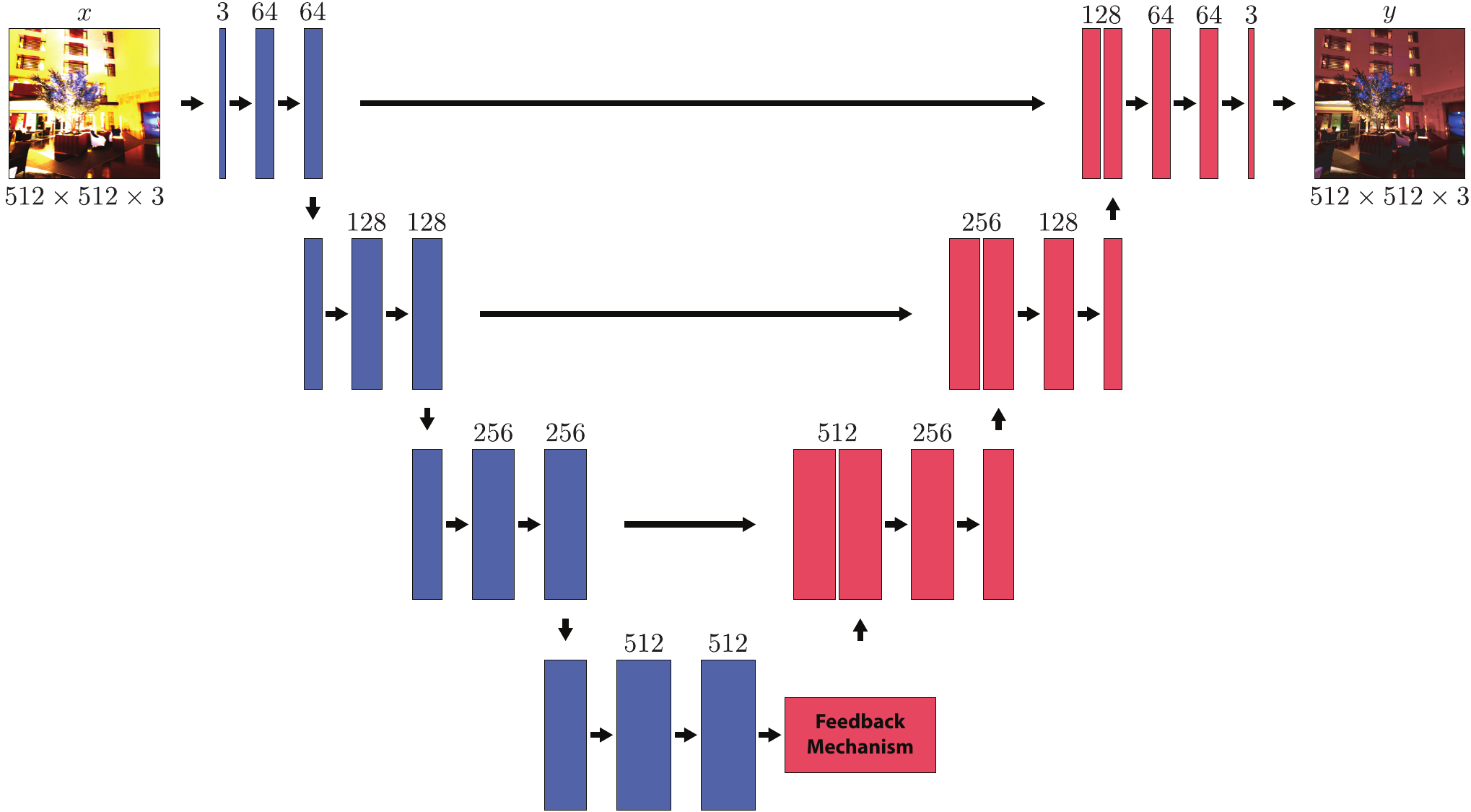}
\caption{Overview of the proposed generators based on our novel feedback based U-Net architecture. Left part (\textcolor{blue}{blue}) is the encoder and right part (\textcolor{red}{red}) is the decoder. The decoder is inside the feedback iteration loop.}
\label{fig:generators}
\end{figure*}

\begin{figure*}[t]
\centering
\includegraphics[width=\linewidth]{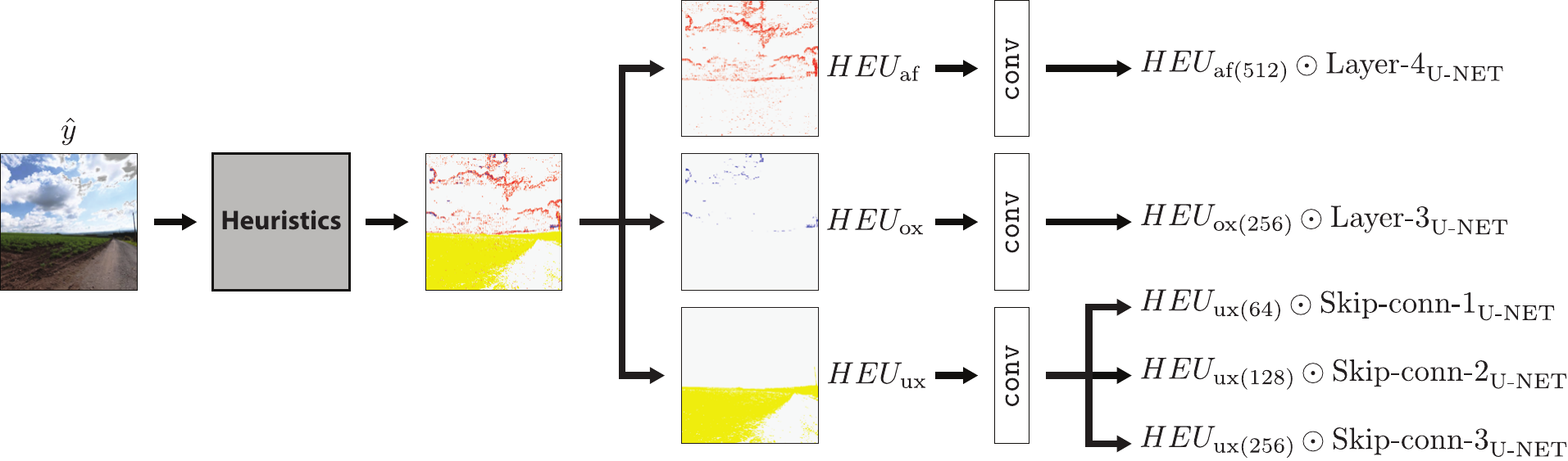}
\caption{Overview of the proposed heuristic-based module. The module outputs the saliency maps for the detected artifacts, overexposed pixels, and underexposed pixels.}
\label{fig:LLM}
\end{figure*}

\subsection{Architecture Modules}
\label{supsubsec:architecture_modules}
\noindent\textbf{Generators.}
The U-Net of each generator consists of $7$ levels.
Each level consists of $2$ \texttt{Conv} layers with $3 \times 3$ kernel followed by a \texttt{ReLU} activation and a \texttt{BN} layer.
The input images of size $512 \times 512$ are first converted into $32$-dimensional features.
The features are doubled in each level of the encoder until reaching size of $512$.
The decoder levels then up-sample the data by performing channel-wise concatenation with the extracted feature from previous levels of the encoder.
The intuition behind using feedback mechanism is to refine the decoder layers of the generator based on the input from encoder layers and the feedback mechanism itself.
The output of each iteration of the feedback mechanism is not only fed into the first level of the decoder block, but also sent back through the hidden state to guide its next input.
Feedback mechanisms perform better than simple feed-forward networks with less trainable parameters~\cite{li2019feedback}.
The feedback network consists of $3$ dilated dense blocks each with dilation rate of $3$.
The dilated blocks are made of $3 \times 3$ \texttt{Conv} layers followed by \texttt{ReLU} activation function.
Each dilated block contains two $1 \times 1$ feature compression layers at the start and end.
Dilation enhances the network's receptive field which helps to capture broader context from the input features without increasing the computational complexity.
Finally, the feedback starts with a $1 \times 1$ \texttt{Conv} and ends with $3 \times 3$ \texttt{Conv} for compression and final processing, respectively.
\cref{fig:generators} depicts and overview of the proposed generators architecture.

\noindent\textbf{Discriminators.}
The discriminators consist of $5$ \texttt{Conv} layers.
We input the reconstructed LDR/HDR along with a real (unpaired) LDR/HDR in the respective discriminator.
The \texttt{Conv} layers have $4 \times 4$ filters and stride of $2$ for the first $3$ layers and $1$ henceforth.
The output of the first layer has features of size $64$, which is doubled in each of the next layers until $512$.
The output of the fifth layer is of size $1$.
The first four layers have \texttt{LeakyReLU} activation with the final layer using a \texttt{Sigmoid} activation for outputting the probabilities of a particular image being real or fake.
The training process is stabilized using \texttt{BN} following all the \texttt{Conv} layers except the first one.

\noindent\textbf{Heuristics.}
We use information from the heuristic-based module to determine whether there are any exposure/artifact issues in the generation.
Specifically, we use OpenCV~\cite{opencv_library} functions to threshold the pixels and generate exposure-aware masks.
Moreover, we use the \texttt{Laplacian} and \texttt{GaussianBlur} functions to generate artifact-aware masks.
The artifact-aware saliency map $HEU_{\text{af}}$ is extracted as feature of size $512$ with $1 \times 1$ \texttt{Conv} and we perform an element-wise multiplication with the output features of the bottleneck layer (\ie, layer $4$) of the U-Net generators $G_Y$ and $G_X$ used in the feedback mechanism.
The exposure-aware saliency map $HEU_{\text{ox}}$ (for overexposure) (used in the generator $G_Y$ only) is extracted as feature of size $256$ with $1 \times 1$ \texttt{Conv} and fused with the output features of layer $3$ henceforth concatenated with the encoder embeddings used in the bottleneck layer.
Finally, $HEU_{\text{ux}}$ (for underexposure) is extracted as feature of sizes $64$, $128$, and $256$, with $1 \times 1$ \texttt{Conv} and fused with the skip connections in each level of the U-Net generator $G_Y$ (\ie, level $1$, $2$, and $3$).
\cref{fig:LLM} depicts the overview of the proposed module.

\noindent\textbf{Encoders.}
The encoders extract embeddings of size $512$ from the input and output domain images.
Given the input size of the generators, we re-project these embeddings to a size of $256$ using $1 \times 1$ \texttt{Conv}.
Then we add the embeddings and concatenate them with the input features of the bottleneck layer (\ie, layer $4$) of our feedback-based U-Net generators.

\subsection{Loss Functions}
\label{supsubsec:loss_functions}
\begin{figure}[h]
\centering
\includegraphics[width=\linewidth]{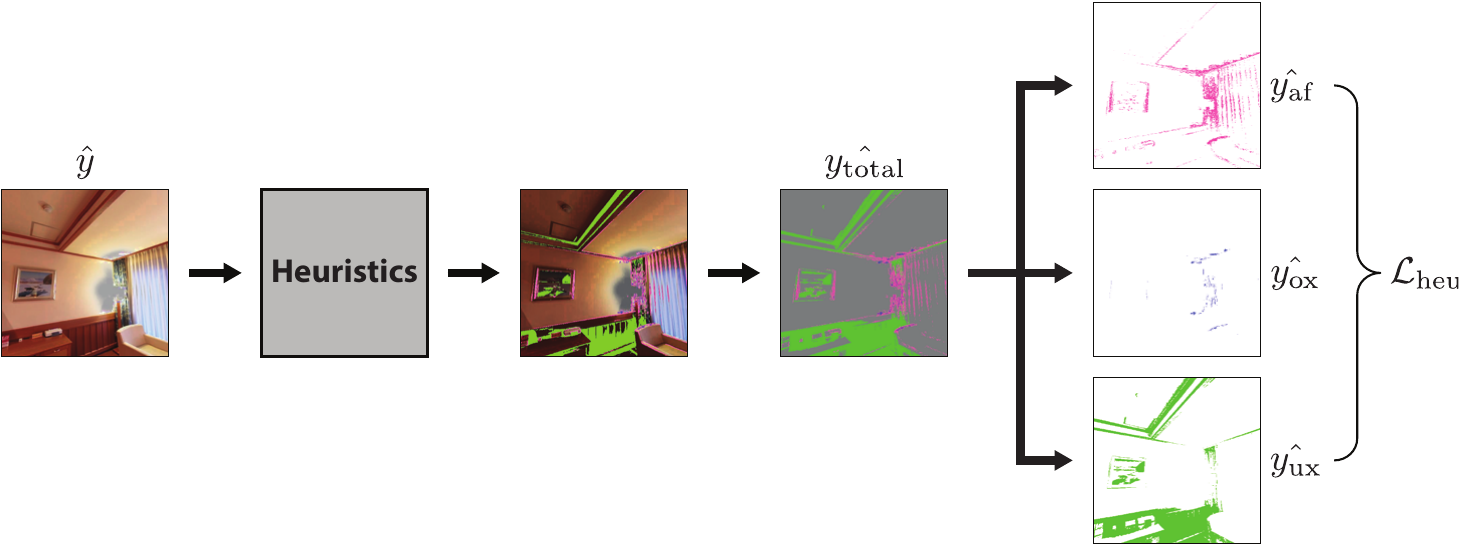}
\caption{Depiction of the heuristic-based loss $\mathcal{L}_{\text{heu}}$. The system outputs the the pixel for the detected artifacts, overexposed pixels, and underexposed pixels.}
\label{fig:llm_loss}
\end{figure}

\begin{figure}[h]
\centering
\includegraphics[width=\linewidth]{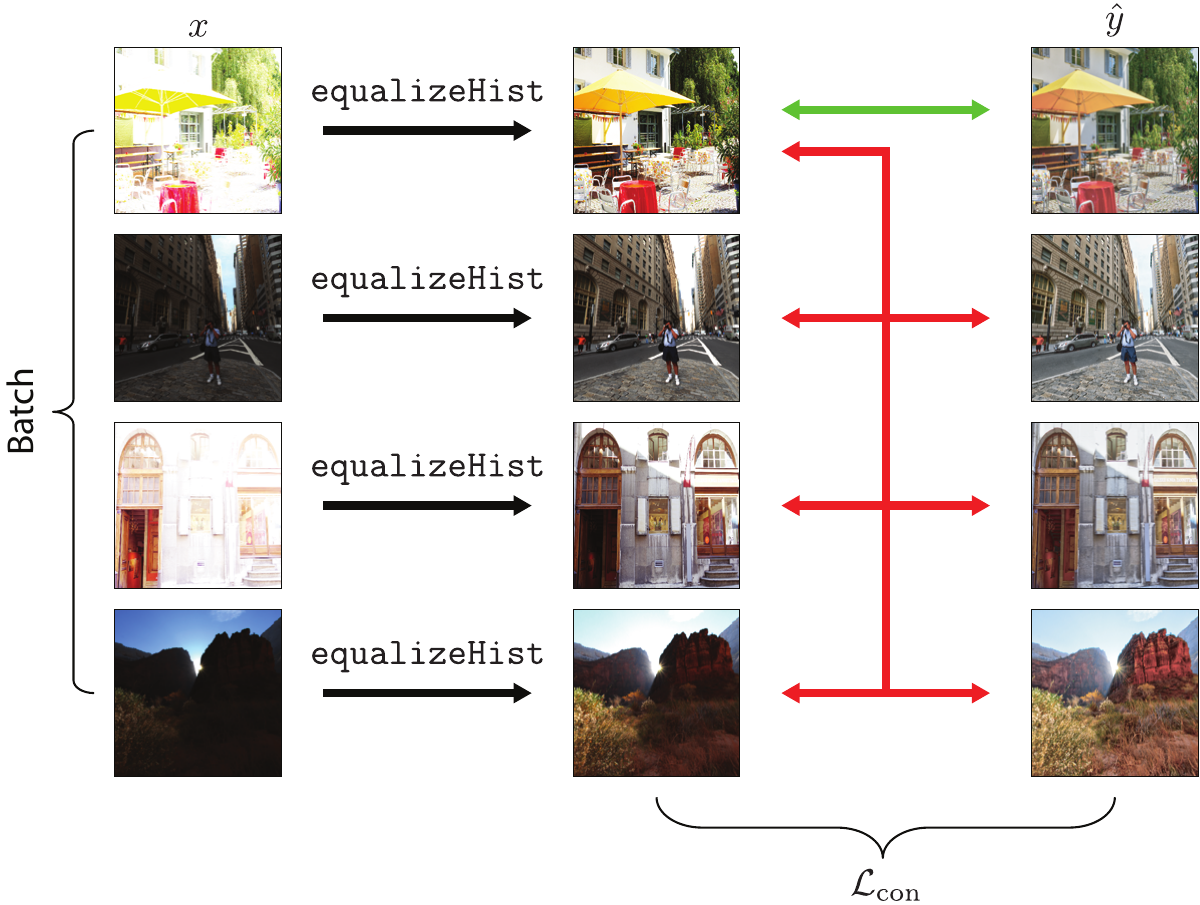}
\caption{Depiction of the contrastive loss $\mathcal{L}_{\text{con}}$. Positive (\textcolor{green}{green}) and negative (\textcolor{red}{red}) pairs in a batch. We use a histogram-equalized version of the LDR processed using the OpenCV function \texttt{equalizeHist}~\cite{opencv_library}.}
\label{fig:contrastive_loss}
\vspace{-5pt}
\end{figure}

\begin{figure}[h]
\centering
\includegraphics[width=\linewidth]{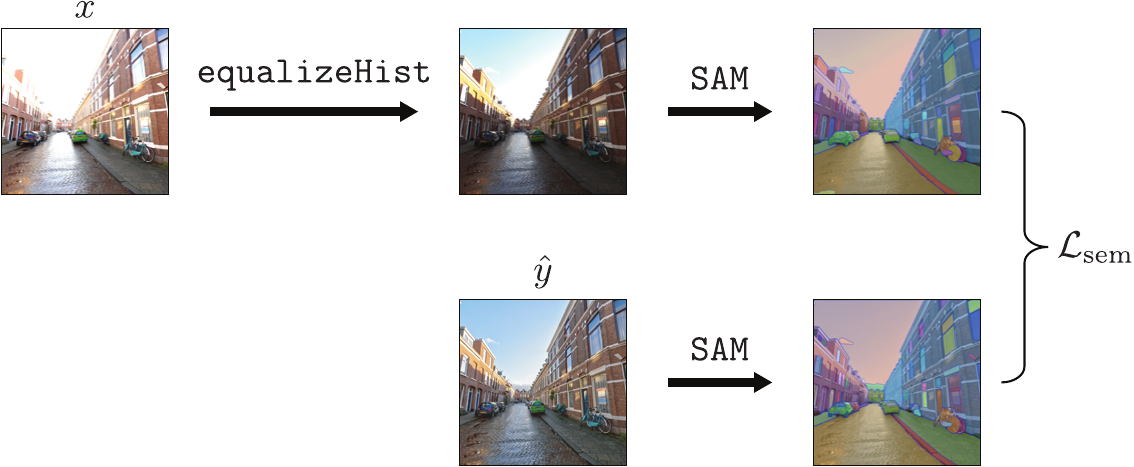}
\caption{Depiction of the semantic segmentation loss $\mathcal{L}_{\text{sem}}$. We use the Segment Anything Model~\cite{kirillov2023segment} to generate segmentation classes in the histogram-equalized LDR and reconstructed tone-mapped HDR images.}
\label{fig:semantic_loss}
\vspace{-5pt}
\end{figure}

\begin{figure}[h]
\centering
\includegraphics[width=\linewidth]{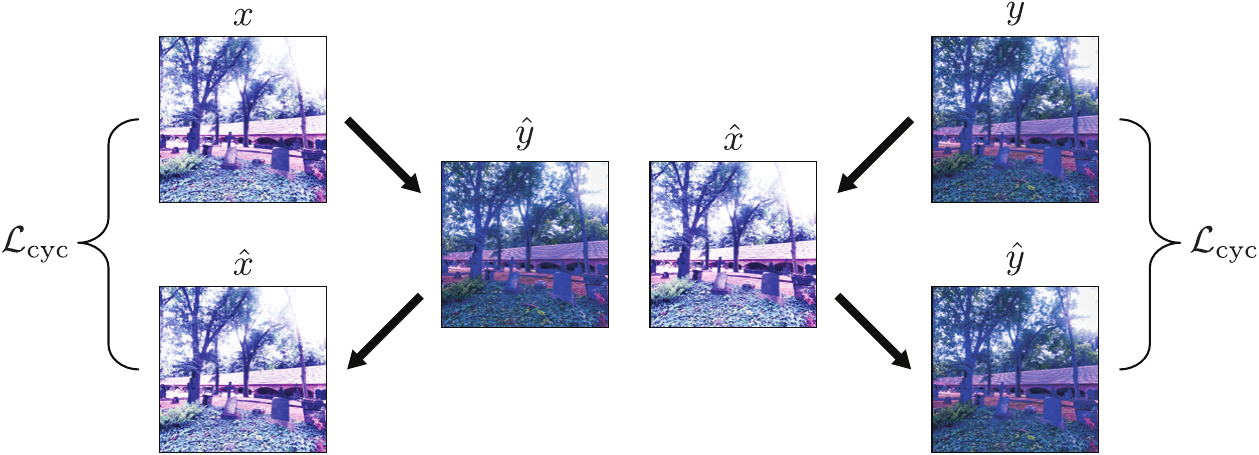}
\caption{Depiction of the cycle consistency loss $\mathcal{L}_{\text{cyc}}$ using an image from the DrTMO~\cite{endo2017deep} dataset.}
\label{fig:cycle_consistency_loss}
\vspace{-5pt}
\end{figure}

\cref{fig:llm_loss,fig:contrastive_loss,fig:semantic_loss,fig:cycle_consistency_loss} illustrate the formulation of the heuristic-based loss, contrastive, semantic segmentation, and cycle consistency functions, respectively.

\section{Results}
\label{supsec:results}
This section complements Sec.~V in the main paper with extensive qualitative results.

\begin{table}[t]
\setlength{\tabcolsep}{2pt}
\centering
\small
\caption{HDR reconstruction results. Comparison with unsupervised learning methods trained on mixed datasets.}
\label{tab:sub_table1}
\begin{tabular}{lcccc}
\toprule[0.5mm]
\textbf{Method} & \textbf{PSNR}$\uparrow$ & \textbf{SSIM}$\uparrow$ & \textbf{LPIPS}$\downarrow$ & \textbf{HDR-VDP-2}$\uparrow$ \\
\midrule[0.25mm]
UPHDR-GAN~\cite{li2022uphdr} & 41.98 & 0.975 & 0.046 & 65.54 \\
GlowGAN-ITM~\cite{wang2023glowgan} & 32.18 & 0.905 & 0.066 & 61.89 \\
\midrule[0.25mm]
\textbf{CycleHDR (Ours)} & \textbf{42.01} & \textbf{0.988} & \textbf{0.023} & \textbf{68.81} \\
\bottomrule[0.5mm]
\end{tabular}
\vspace{-10pt}
\end{table}

\subsection{HDR Reconstruction}
\label{supsubsec:hdr_reconstruction}
\cref{tab:sub_table1} provides the results from an additional mixed data experiment.
The train set used in this experiment is: 1) The HDR images from DrTMO~\cite{endo2017deep}, Kalantari~\cite{kalantari2017deep}, HDR-Eye~\cite{korshunov2014crowdsourcing}, LDR-HDR Pair~\cite{jang2020dynamic}, and GTA-HDR~\cite{barua2024gta} ($1500$ random samples), \ie, $2854$ HDR images; and 2) LDR images from GTA-HDR~\cite{barua2024gta} ($2000$ random samples) and $50\%$ of the LDR images in DrTMO, Kalantari, HDR-Eye, and LDR-HDR Pair, \ie, $2677$ LDR images.
In this dataset, a total of $1177$ \{LDR,HDR\} pairs overlap.
Given that the only other methods that support this type of unpaired train data are UPHDR-GAN~\cite{li2022uphdr} and GlowGAN-ITM~\cite{wang2023glowgan}, we perform a direct comparison with these two approaches.
The results demonstrate that our method again outperforms UPHDR-GAN and GlowGAN-ITM on all metrics.

\begin{figure*}[t]
\centering
\includegraphics[width=\linewidth]{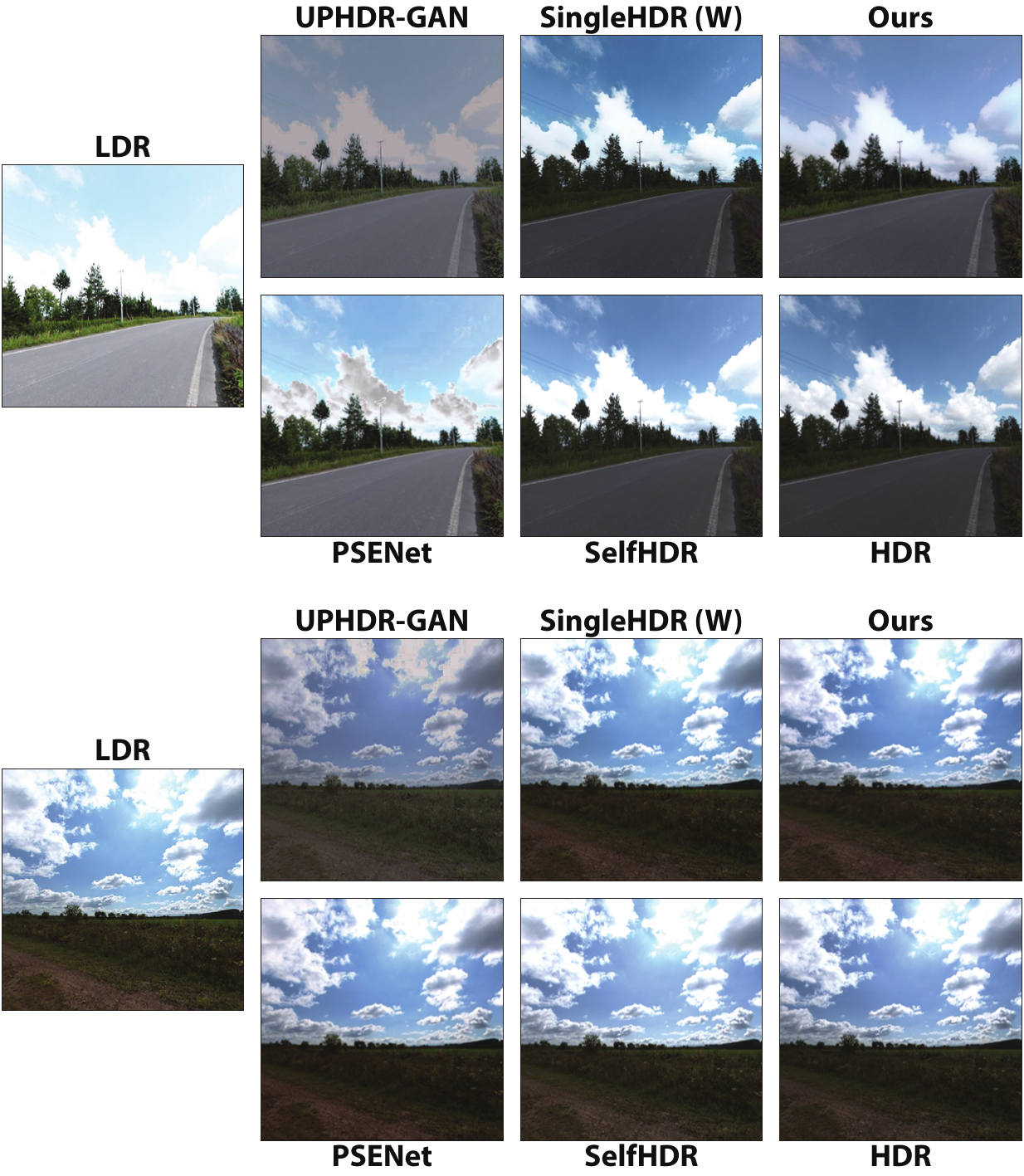}
\caption{Examples of HDR images reconstructed with our method and recent state-of-the-art.}
\label{fig:results2}
\end{figure*}

\begin{figure*}[t]
\centering
\includegraphics[width=\linewidth]{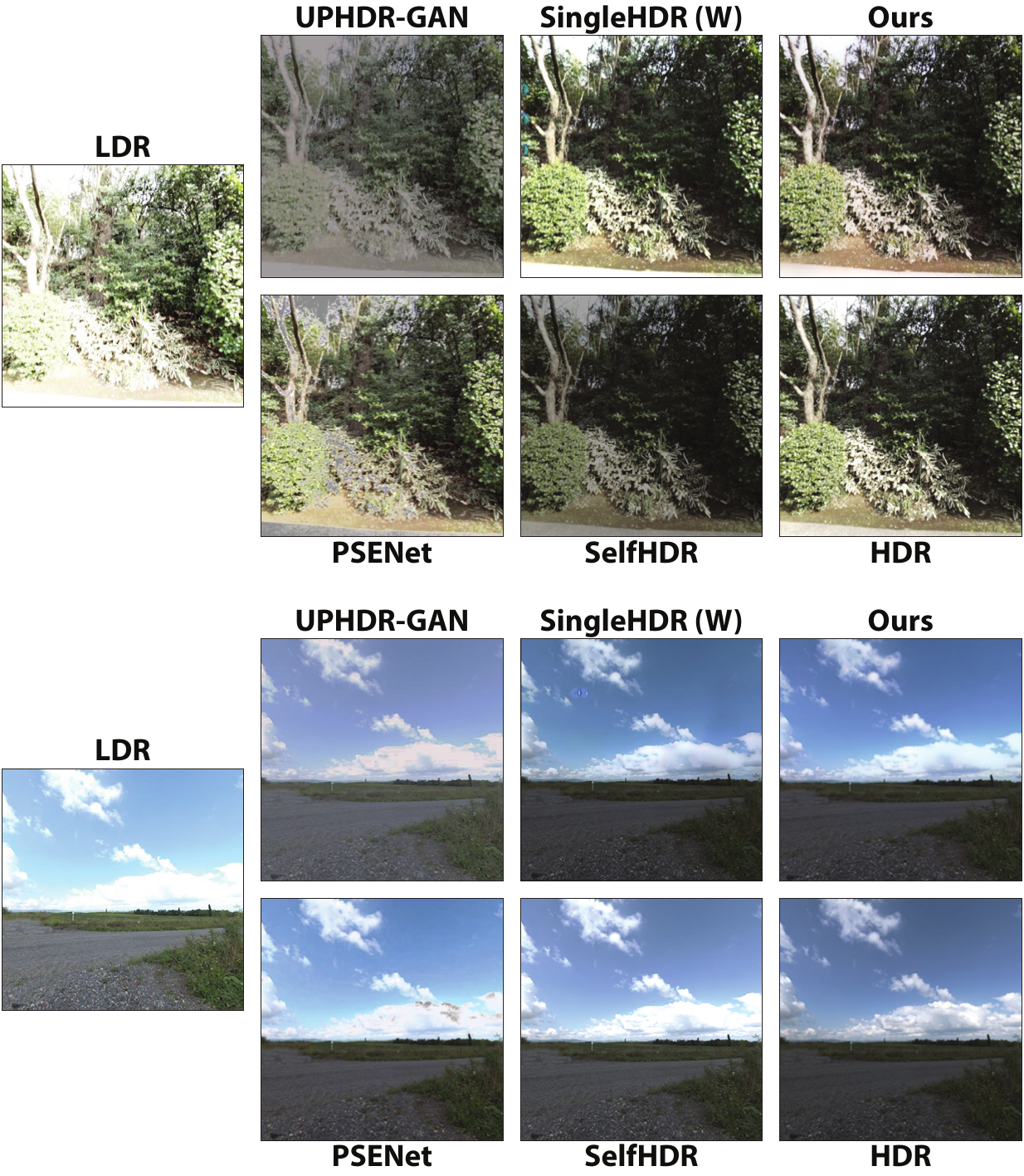}
\caption{Examples of HDR images reconstructed with our method and recent state-of-the-art.}
\label{fig:results3}
\end{figure*}

\begin{figure*}[t]
\centering
\includegraphics[width=\linewidth]{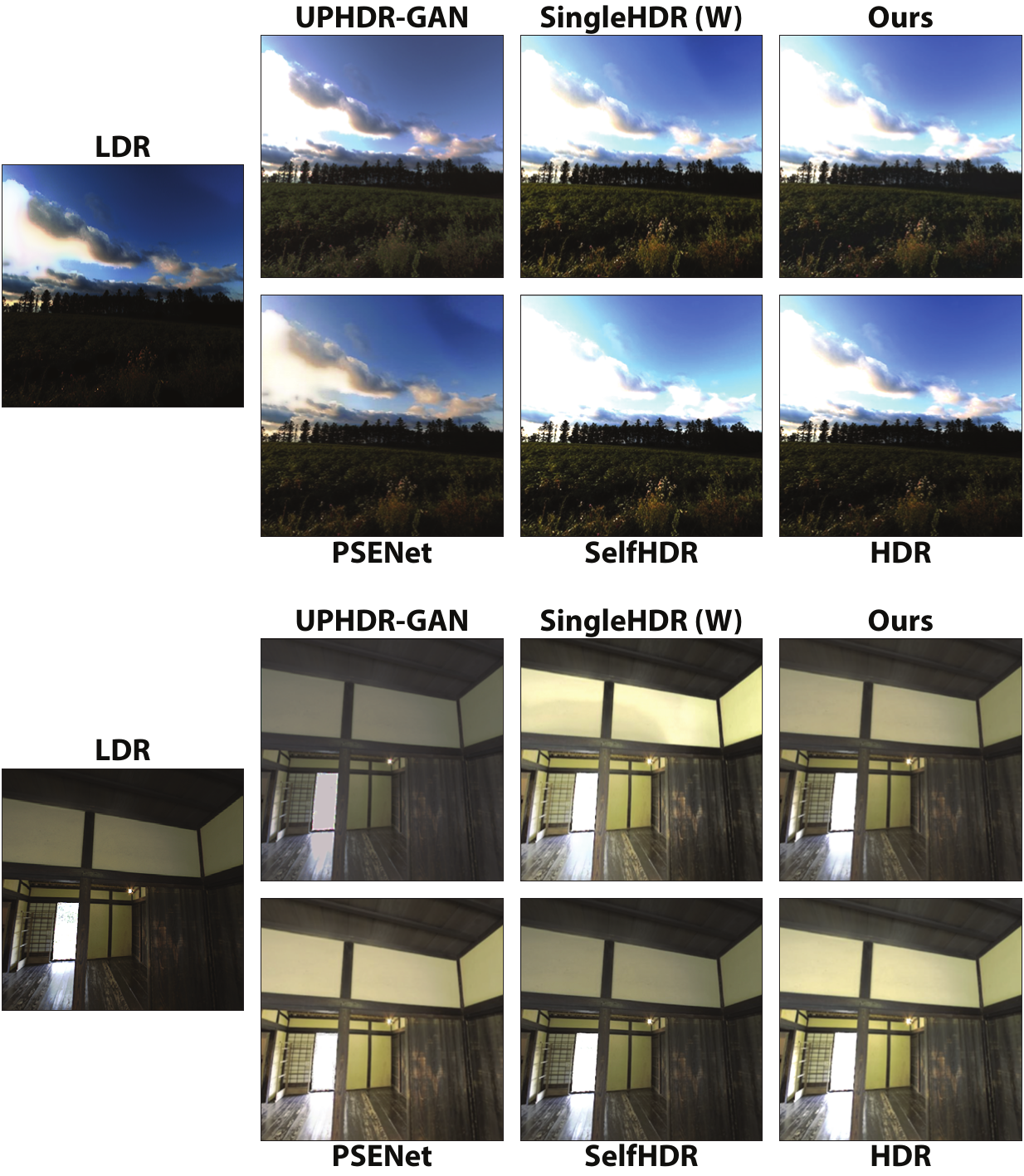}
\caption{Examples of HDR images reconstructed with our method and recent state-of-the-art.}
\label{fig:results4}
\end{figure*}

\begin{figure*}[t]
\centering
\includegraphics[width=\linewidth]{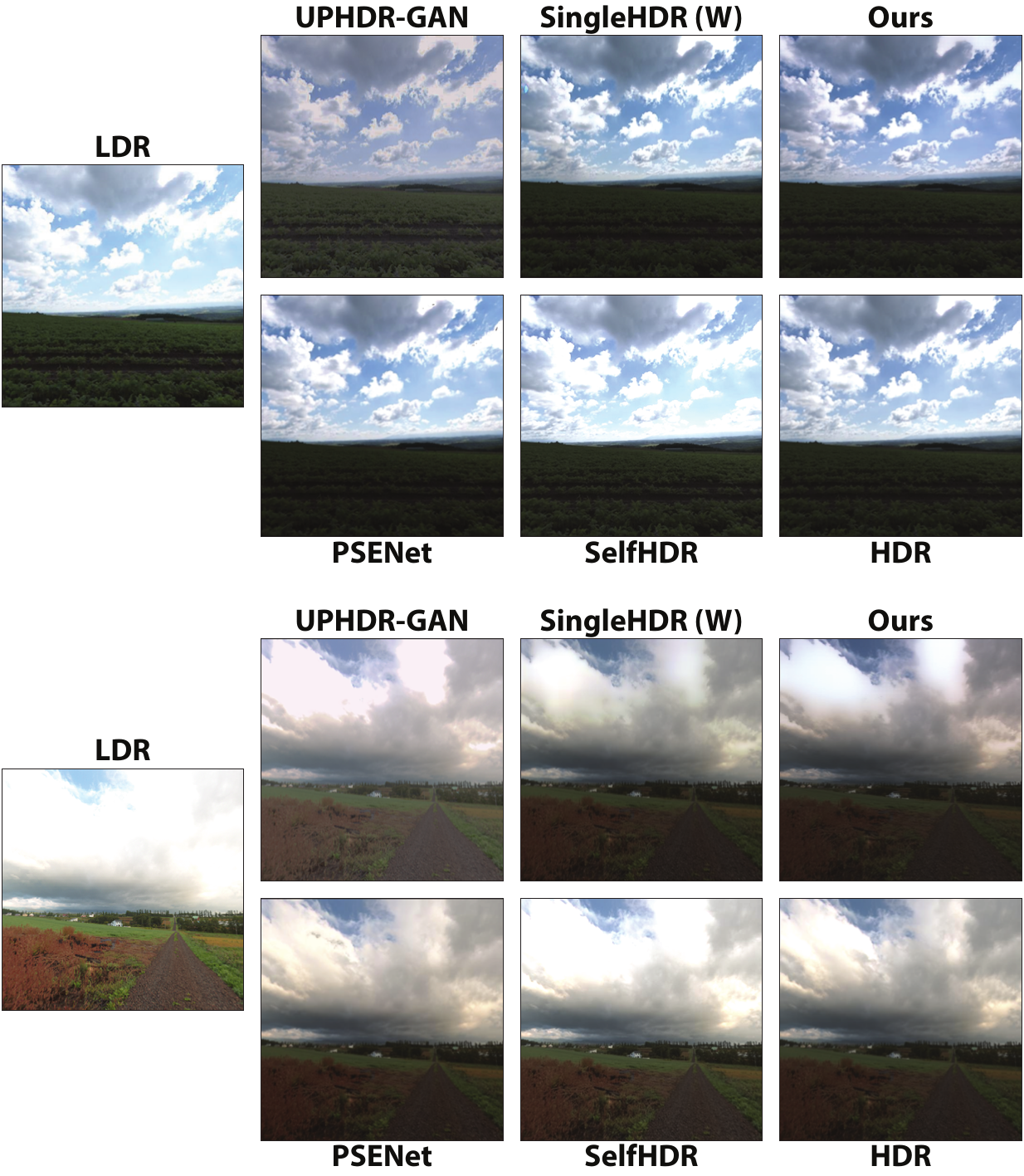}
\caption{Examples of HDR images reconstructed with our method and recent state-of-the-art.}
\label{fig:results5}
\end{figure*}

\begin{figure*}[t]
\centering
\includegraphics[width=\linewidth]{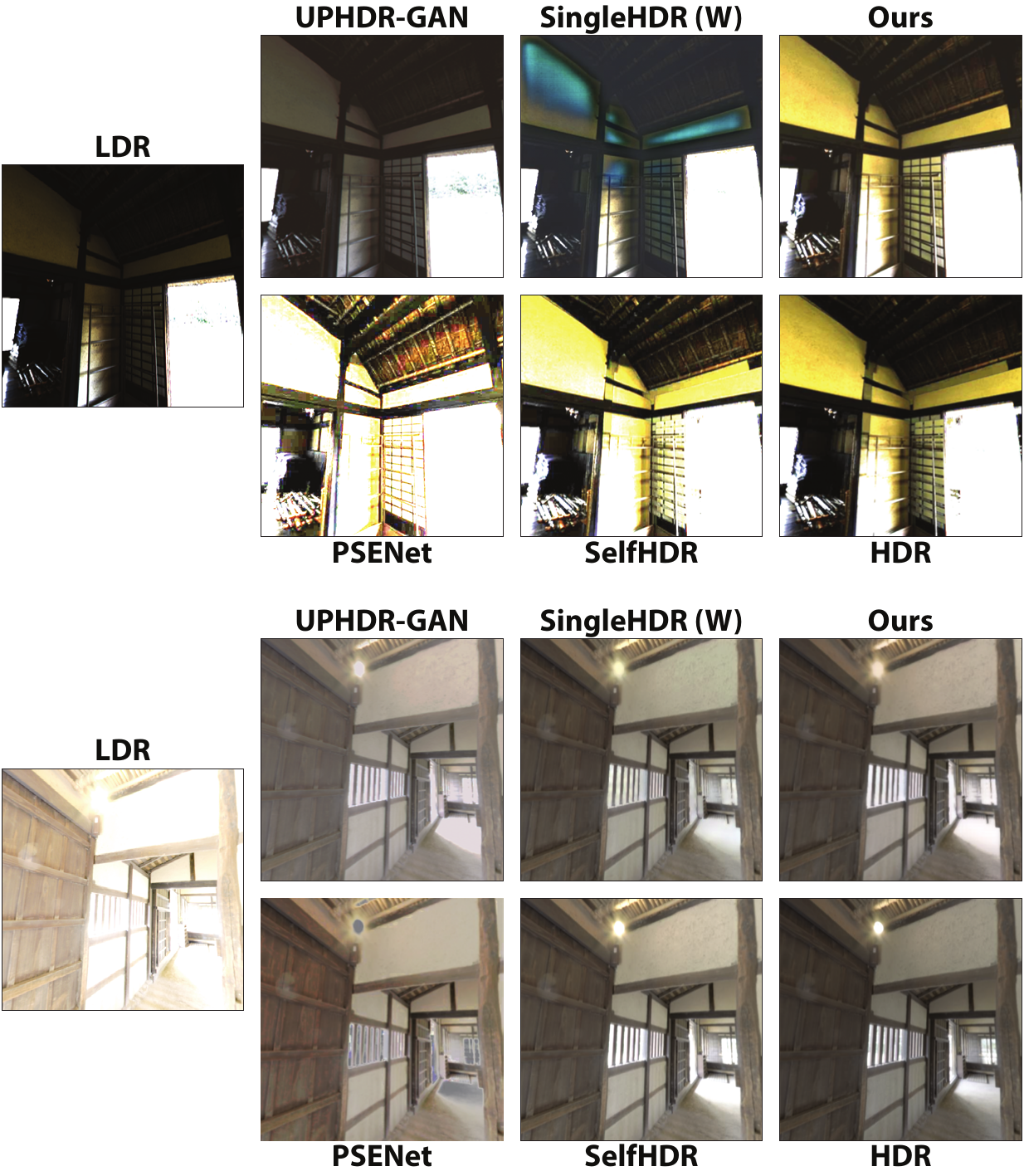}
\caption{Examples of HDR images reconstructed with our method and recent state-of-the-art.}
\label{fig:results6}
\end{figure*}

\begin{figure*}[t]
\centering
\includegraphics[width=\linewidth]{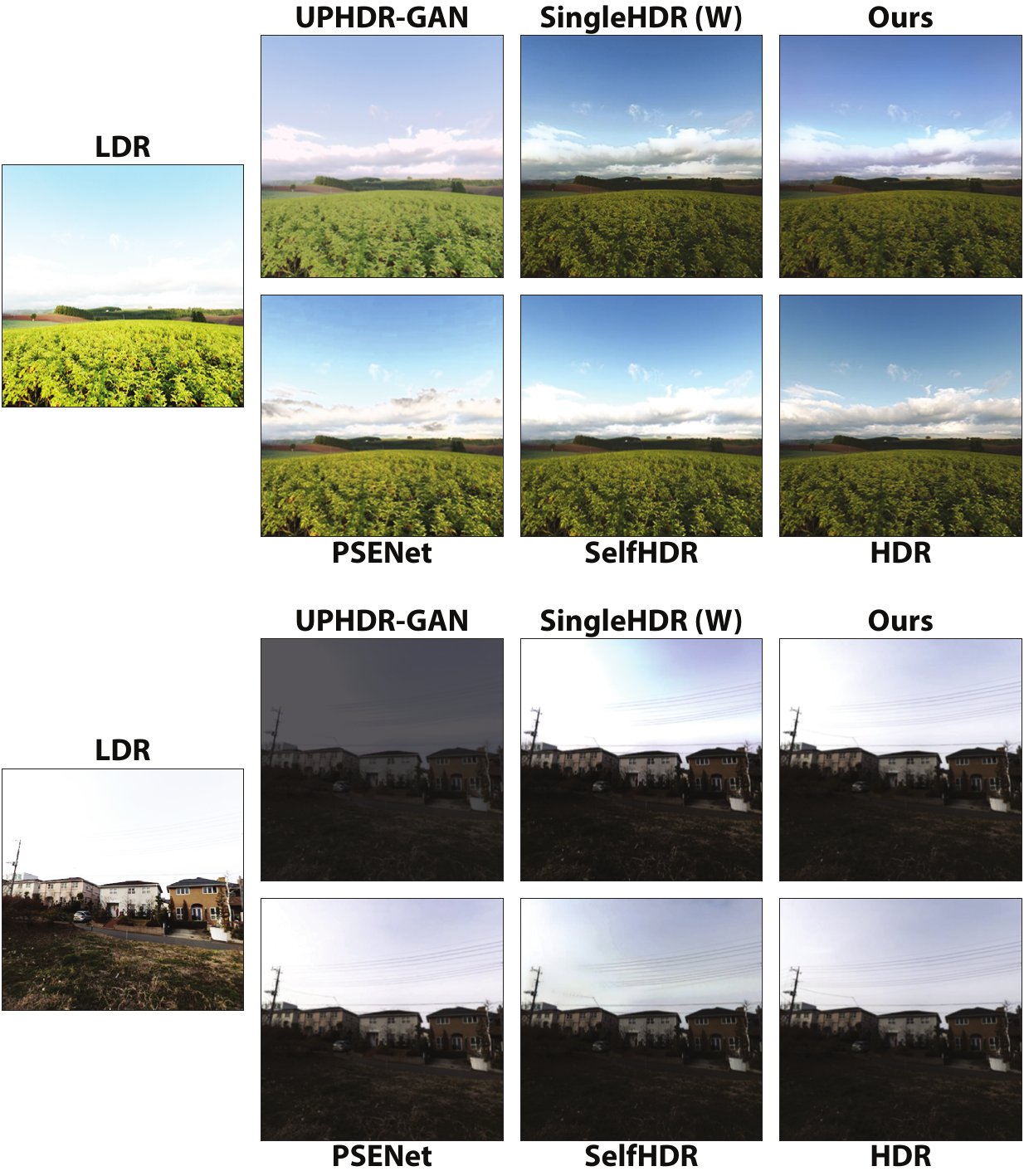}
\caption{Examples of HDR images reconstructed with our method and recent state-of-the-art.}
\label{fig:results7}
\end{figure*}

\cref{fig:results2,fig:results3,fig:results4,fig:results5,fig:results6,fig:results7} illustrate the quality of the HDR images reconstructed with our method.
Our method mitigates artifacts such that the resulting HDR closely resembles the ground truth HDR.
The finer textures, color hues, and shades are well preserved compared to the other methods.
The granular level details in the bright and dark regions are also well reconstructed.

\subsection{LDR Reconstruction}
\label{supsubsec:ldr_reconstruction}
\cref{tab:tonemap} provides the results from an additional LDR reconstruction experiment.
In this case, our method is trained with the data specified in the previous section.
Similar to the LDR reconstruction experiment in the main paper, we compare the model with the state-of-the-art tone-mapping operators $\mu$-law~\cite{jinno2011mu}, Reinhard's~\cite{reinhard2023photographic},  Photomatix~\cite{photomatix}, GlowGAN-prior~\cite{wang2023glowgan}, Unpaired-TM~\cite{vinker2021unpaired}, DeepTMO~\cite{rana2019deep}, and Zero-shot-TM~\cite{zhu2024zero} for comparison.
The results again demonstrate that our method outperforms the other approaches in terms of SSIM and LPIPS.
In terms of PSNR, it is third best after DeepTMO and Photomatix.

\begin{table}[t]
\setlength{\tabcolsep}{2pt}
\centering
\small
\caption{LDR reconstruction results. Comparison with the state-of-the-art tone-mapping operators and GlowGAN-prior~\cite{wang2023glowgan}.}
\label{tab:tonemap}
\begin{tabular}{lccc}
\toprule[0.5mm]
\textbf{Method} & \textbf{PSNR}$\uparrow$ & \textbf{SSIM}$\uparrow$ & \textbf{LPIPS}$\downarrow$ \\
\midrule[0.25mm]
$\mu$-law operator~\cite{jinno2011mu} & 28.68 & 0.891 & 0.311 \\
Reinhard's operator~\cite{reinhard2023photographic} & 30.12 & 0.903 & 0.342 \\
Photomatix~\cite{photomatix} & 32.11 & 0.922 & 0.136 \\
DeepTMO~\cite{rana2019deep} & \textbf{32.13} & 0.923 & 0.133 \\
Unpaired-TM~\cite{vinker2021unpaired} & 31.62 & 0.925 & 0.117 \\
Zero-shot-TM~\cite{zhu2024zero} & 31.71 & 0.929 & 0.110 \\
GlowGAN-prior~\cite{wang2023glowgan} & 31.91 & 0.931 & 0.112 \\
\midrule[0.25mm]
\textbf{CycleHDR (Ours)} & 32.05 & \textbf{0.939} & \textbf{0.095} \\
\bottomrule[0.5mm]
\end{tabular}
\vspace{-10pt}
\end{table}

\end{document}